\documentclass[a4paper,fleqn]{cas-dc}

\usepackage[authoryear]{natbib}

\usepackage{soul,color}
\usepackage{subfigure}
\usepackage{commath}
\usepackage{caption}
\usepackage{url}
\usepackage{tikz}
\newcommand{\Fone}{F\textsubscript{1}}

\def\tsc#1{\csdef{#1}{\textsc{\lowercase{#1}}\xspace}}
\tsc{WGM}
\tsc{QE}
\tsc{EP}
\tsc{PMS}
\tsc{BEC}
\tsc{DE}

\begin{document}
\let\WriteBookmarks\relax
\def\floatpagepagefraction{1}
\def\textpagefraction{.001}
\shorttitle{Determination of building flood risk maps from LiDAR mobile mapping data}
\shortauthors{Feng et~al.}

\title [mode = title]{Determination of building flood risk maps from LiDAR mobile mapping data}




\author[1]{Yu Feng}[orcid=0000-0001-5110-5564]
\cormark[1]
\ead{yu.feng@ikg.uni-hannover.de}

\author[1]{Qing Xiao}

\author[1]{Claus Brenner}

\author[2,3]{Aaron Peche}

\author[4,5]{Juntao Yang}

\author[1]{Udo Feuerhake}

\author[1]{Monika Sester}

\address[1]{Institute of Cartography and Geoinformatics, Leibniz University Hannover, Appelstra\ss{}e 9a,
30167 Hannover, Germany}
\address[2]{bpi Hannover–Beratende Ingenieure, Mengendamm 16d, 30177 Hanover, Germany}
\address[3]{Federal Institute for Geosciences and Natural Resources, 30655 Hannover, Germany}
\address[4]{College of Geomatics, Shandong University of Science and Technology, Qingdao 266590, China}
\address[5]{School of Land Science and Technology, China University of Geosciences, No. 29 Xueyuan Road, Haidian District, Beijing 100083, China}

\cortext[cor1]{Corresponding author}












\begin{abstract}
With increasing urbanization, flooding is a major challenge for many cities today. Based on forecast precipitation, topography, and pipe networks, flood simulations can provide early warnings for areas and buildings at risk of flooding. Basement windows, doors, and underground garage entrances are common places where floodwater can flow into a building. Some buildings have been prepared or designed considering the threat of flooding, but others have not. Therefore, knowing the heights of these facade openings helps to identify places that are more susceptible to water ingress. However, such data is not yet readily available in most cities. Traditional surveying of the desired targets may be used, but this is a very time-consuming and laborious process. Instead, mobile mapping using LiDAR (light detection and ranging) is an efficient tool to obtain a large amount of high-density 3D measurement data. To use this method, it is required to extract the desired facade openings from the data in a fully automatic manner.

This research presents a new process for the extraction of windows and doors from LiDAR mobile mapping data. Deep learning object detection models are trained to identify these objects. Usually, this requires to provide large amounts of manual annotations. In this paper, we mitigate this problem by leveraging a rule-based method. In a first step, the rule-based method is used to generate pseudo-labels. A semi-supervised learning strategy is then applied with three different levels of supervision. The results show that using only automatically generated pseudo-labels, the learning-based model outperforms the rule-based approach by 14.6\% in terms of \Fone-score. After five hours of human supervision, it is possible to improve the model by another 6.2\%.

By comparing the detected facade openings' heights with the predicted water levels from a flood simulation model, a map can be produced which assigns per-building flood risk levels. Thus, our research provides a new geographic information layer for fine-grained urban emergency response. This information can be combined with flood forecasting to provide a more targeted disaster prevention guide for the city's infrastructure and residential buildings. To the best of our knowledge, this work is the first attempt to achieve such a large scale, fine-grained building flood risk mapping.
\end{abstract}


\maketitle

\section{Introduction}
Floods are the most frequent natural disaster, causing many deaths, injuries, and property damages.
With rapid urbanization, more flood-prone areas have been developed, exposing more buildings to flood risk \citep{rozer2021current}.

With improved computational resources and more efficient numerical methods and algorithms, the use of complex coupled pipe-overland flow models for flood forecasting is increasing \citep{vojinovic2009,martins2018}. In recent studies, the development of surrogate models improved the efficiency of flood models even further \citep{bermudez2018development,berkhahn2019ensemble}.
With an artificial-neural-network-based inundation model, the required time for flood simulation is greatly reduced to the order of real-time \citep{rozer2021impact}.
It is expected that flood simulations will be used more actively for emergency response.

By overlaying with building footprint data, flood simulations can highlight buildings that intersect with the estimated flood extent \citep{correia1998coupling, ernst2010micro}. However, the actual threat may vary from building to building.
Flood water mostly enters the building through doors, windows, and underground garage entrances \citep{spekkers2017comparative}.
Many buildings have obvious protective measures against the potential threat of flooding, such as adding steps to doors and raising the height of basement windows. 
It is only when flooding is severe enough that these targets, i.e. facade openings, will be affected.
Without considering information on individual buildings, flood simulations may identify a large number of buildings, many of which are not actually at risk.

Therefore, it is important to estimate in more detail the likelihood of water entering the building and where it may enter. 
This helps to identify areas that may be in more need of emergency resources, such as sandbags and movable barriers to prevent floodwater ingress, or life rafts and emergency crews to check for trapped people during the flood. 
Residents can also receive guidance on where these barriers should be prioritized for allocation.
With this, targeted prevention measures can be applied to achieve a fine-grained emergency response.
Consequently, more accurate per-building information is needed, especially with respect to the heights of the doors and windows, in order to determine the threat.

3D city models are the ideal surveying product to cope with such scenarios. The models are often categorized into four levels of detail (LoD). LoD1 are buildings in rectangular blocks. Roofs are added for LoD2, and LoD3 adds windows and doors on facades. Finally, LoD4 combines the modeling of indoor spaces \citep{biljecki2016improved}. This means that products of LoD3 and LoD4 would be suitable to achieve the desired fine-grained flood risk mapping. However, while LoD1 and LoD2 models are currently available for many cities, LoD3 and LoD4 models are not commonly available \citep{delft2020datasets}.
This may be due to the fact that the modeling of windows and doors on the facade is one of the challenging, and thus expensive, steps of 3D city modeling.
Obtaining such fine-grained 3D models of cities requires high-precision measurements, sophisticated algorithms, and also laborious refinement.

LiDAR mobile mapping systems are effective tools for obtaining large-scale urban 3D measurements.
The acquired point cloud data have a wide range of applications in surveying and mapping, and also for autonomous driving. It is an important data source for producing 3D city models of a higher level of detail \citep{wang2018lidar,xia2020geometric}, such as LoD3 in \cite{wen2019accurate} and LoD4 using indoor modeling in \cite{buyuksalih20193d}.
The buildings' facades can often be measured with a very high point density, especially in the areas adjacent to the ground.
This provides suitable quality data for extracting the windows and doors of the facade.
Thus, in areas where detailed 3D city models are not available, LiDAR mobile mapping data can be acquired and then used to extract targets of interest.

This paper aims to extract windows and doors on building facades including their heights.
By combining this with the flood simulation results, windows and doors exposed to flood risk can be automatically identified.
The flood risk map generated from our proposed process has a great potential to serve as a new geographic information layer for city emergency management.
Besides this benefit for disaster management, the proposed method introduces a novel semi-supervised learning strategy that can effectively reduce the need for data annotation when applying deep learning models. 

The remainder of this paper is organized as follows: Section \ref{sec:related_work} introduces related work. In Section \ref{sec:methodology}, the proposed process is presented, which consists of five steps: extraction of building facades, rule-based object detection using scan lines, deep-learning-based object detection using semi-supervised learning, post-processing, and mapping of building flood risks.
Section \ref{sec:experiments} presents an experiment on the mobile mapping LiDAR point cloud data collected in Hildes\-heim, Germany. 
The methods to extract windows and doors are compared qualitatively, and are evaluated quantitatively based on a human-annotated dataset.
Furthermore, locations with high flood risk are identified based on flood simulation results. High flood risk maps are generated and verified by examining the point cloud of individual locations. A conclusion and an outlook are given in the last section.

\section{Related work}
\label{sec:related_work}

Methods for window detection from LiDAR point clouds can be categorized into two groups. One is based on edge detection, and the other is based on hole detection \citep{xia2019faccade}.

The edge-based approach first detects edge points, from which the window shape is subsequently reconstructed.
The reliable detection of boundary points is an essential step.
\cite{wang2011window, wang2012method} pre-process the facade point cloud into voxels and inspect each point with its top, bottom, left, and right neighboring voxels. Once there are points in an arbitrary direction and no points in the opposite direction, this location is possibly part of the boundary. 
\cite{nguatem2014localization} applied the boundary estimation algorithm of \cite{rusu2007towards}.
FPFH (fast point feature histograms) \citep{rusu2009fast} were used to identify edge points based on point saliency in \cite{hao2019saliency}.
There are also different strategies to reconstruct window shapes. Template matching was used in \cite{nguatem2014localization}, where three window templates were considered. Based on the assumption that windows are arranged in a regular grid pattern, \cite{wang2012method,peethambaran2017enhancing} identified peaks and valleys from horizontal and vertical edge point distributions. Windows are first detected as the intersection points of peaks in both directions. Then, windows were reconstructed using the nearby edge points.
Windows on the ground floor are often ignored, e.g., in \cite{wang2011window}, because they mostly do not meet this assumption.

Hole-based methods detect the holes in the facades based on the assumption that the windows are complete holes without any reflection.
\cite{pu2009knowledge} and \cite{boulaassal2009automatic} identify the edge points using triangulation. The points connected with long edges are regarded as edge points and then grouped using clustering. Rectangular shapes are matched to the detected clusters. The varying point density of mobile mapping LiDAR data could be a challenge for such methods.
\cite{zolanvari2016slicing, zolanvari2018three} proposed a slicing method to scan the point cloud in horizontal and vertical directions. The slices' endpoints are used to identify the boundary points of opening areas on the facades point cloud.
In addition, \cite{xia2019faccade} directly identified holes on the facade segment based on the connectivity grid provided by the laser scanner. Connected-component analysis \citep{mesolongitis2012detection} was used, where regions without projected points are marked as potential window segments. Small and irregular segments were filtered using rules. 
However, it remains a challenge for the hole detection-based methods to cope with scenarios where there are facades with varying point densities and missing points due to occlusions, which usually cannot be avoided due to the principles of mobile mapping LiDAR.

\begin{figure*}
    \centering
    \includegraphics[clip, trim=0 70 0 0, width=\textwidth]{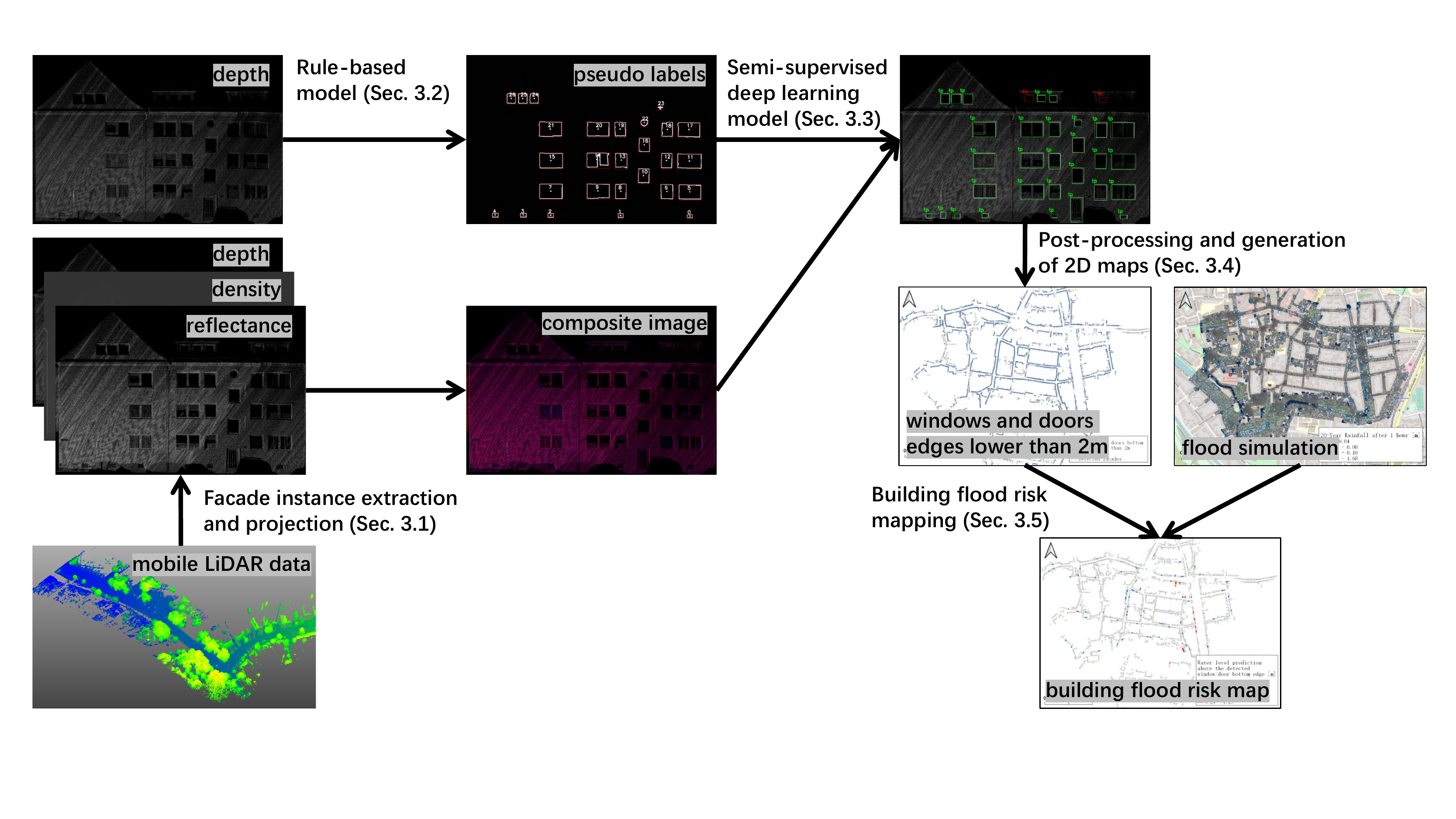}\\
    \caption{Overview of the proposed process for building flood risk mapping.}
    \label{fig:process_overview}
\end{figure*}

In the large-scale mobile mapping LiDAR data acquisition presented in this work, more complicated situations are considered.
First, windows are not always complete holes, due to objects like curtains, items on windowsills, and window stickers. This leads to low-quality results when applying simple approaches such as the slicing method.
Second, windows and doors on the ground floor are most important for our application, and therefore they must not be neglected.
Third, facades are acquired with very different point densities, e.g., dense near the ground and sparse near the roof. There are also some facades scanned from a distance,  which have a low point density. As this is unavoidable, more robust models are required that can deal with this, and are able to still detect as many high-quality facade openings as possible. 
In general, rule-based methods involve many parameters defining the shape, the size, or the degree of protrusion and depression of windows. These parameters perform well in some scenarios but may fail in others.
Deep learning based methods have been introduced to facade modeling in recent years and have achieved significantly better performance compared to rule-based methods. 
However, most of them are color image-based, e.g.,
\cite{schmitz2016convolutional, liu2017deepfacade, liu2020deepfacade, kong2020enhanced}. Deep learning methods have rarely been applied to a similar extent for point cloud-based facade modeling.
In this work, we consider the detection based on the projected 2D images of the facade point cloud. Detection on such projected images was also applied in many rule-based methods, e.g., \cite{ripperda2006reconstruction, mesolongitis2012detection, xia2019faccade}.

With regard to the application scenarios, private damage-reducing measures nowadays play an important role in flood management. The residents are often informed of the flood risk. However, their actions depend heavily on their personal assessment \citep{kreibich2015review}. Flood simulation analysis can enhance their awareness of flood risk \citep{sanders2020collaborative}.
\cite{mazzorana2014physical} emphasized that the geometry characteristics of individual buildings should be considered in vulnerability analysis. 
The positions of the facade openings are among the important indicators.
A detailed BIM (Building Information Model) of a single house has been considered in combination with flood simulation results for building damage assessment \citep{amirebrahimi2016bim}. However, finely crafted BIMs are often costly and difficult to obtain on a large scale.
As for the flood risk assessment on a large scale, resource-consuming fieldwork \citep{fedeski2007urban} has to be done to collect vulnerability information for individual buildings, which includes the facade opening heights.


To the best of our knowledge, there have not been substantial research efforts in the past to integrate the facade openings detection with the application of identifying flood risk locations for buildings.
While \cite{van2019extracting} described a similar idea of using detected doors and windows on panoramic images and mobile mapping LiDAR data to estimate the impact of the flood on buildings, no detailed experiments to test its feasibility were reported. 
Therefore, we think our proposed process is a first attempt to implement the entire process and combine it with the flood simulation results to identify high flood-risk locations for buildings.


\section{Methodology}
\label{sec:methodology}

\begin{figure*}
    \centering
    \subfigure{
    \begin{minipage}[t]{.48\linewidth}
    \centering
    \includegraphics[width=\textwidth]{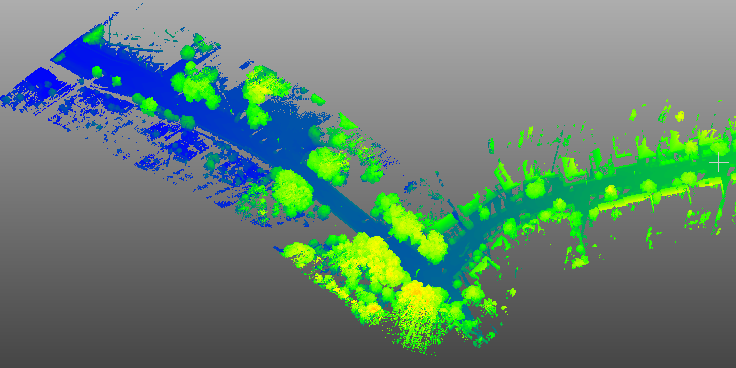}\\(a) 3D point cloud
    \end{minipage}%
    }%
    \subfigure{
    \begin{minipage}[t]{.48\linewidth}
    \centering
    \includegraphics[width=\textwidth]{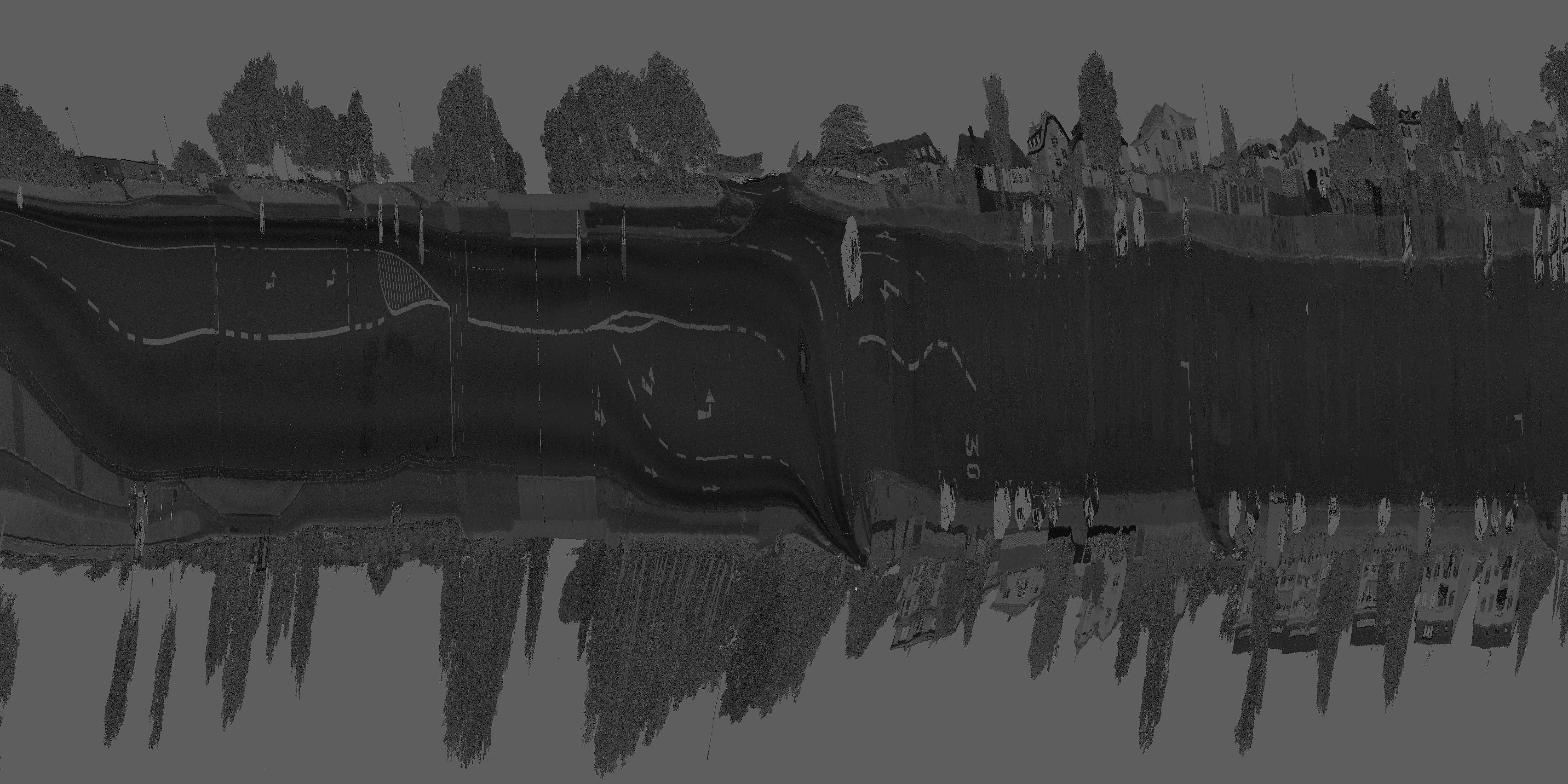}\\ (b) scan strip image of reflectance
    \end{minipage}%
    }%
    \vspace{1mm}
    \subfigure{
    \begin{minipage}[t]{.48\linewidth}
    \centering
    \includegraphics[width=\textwidth]{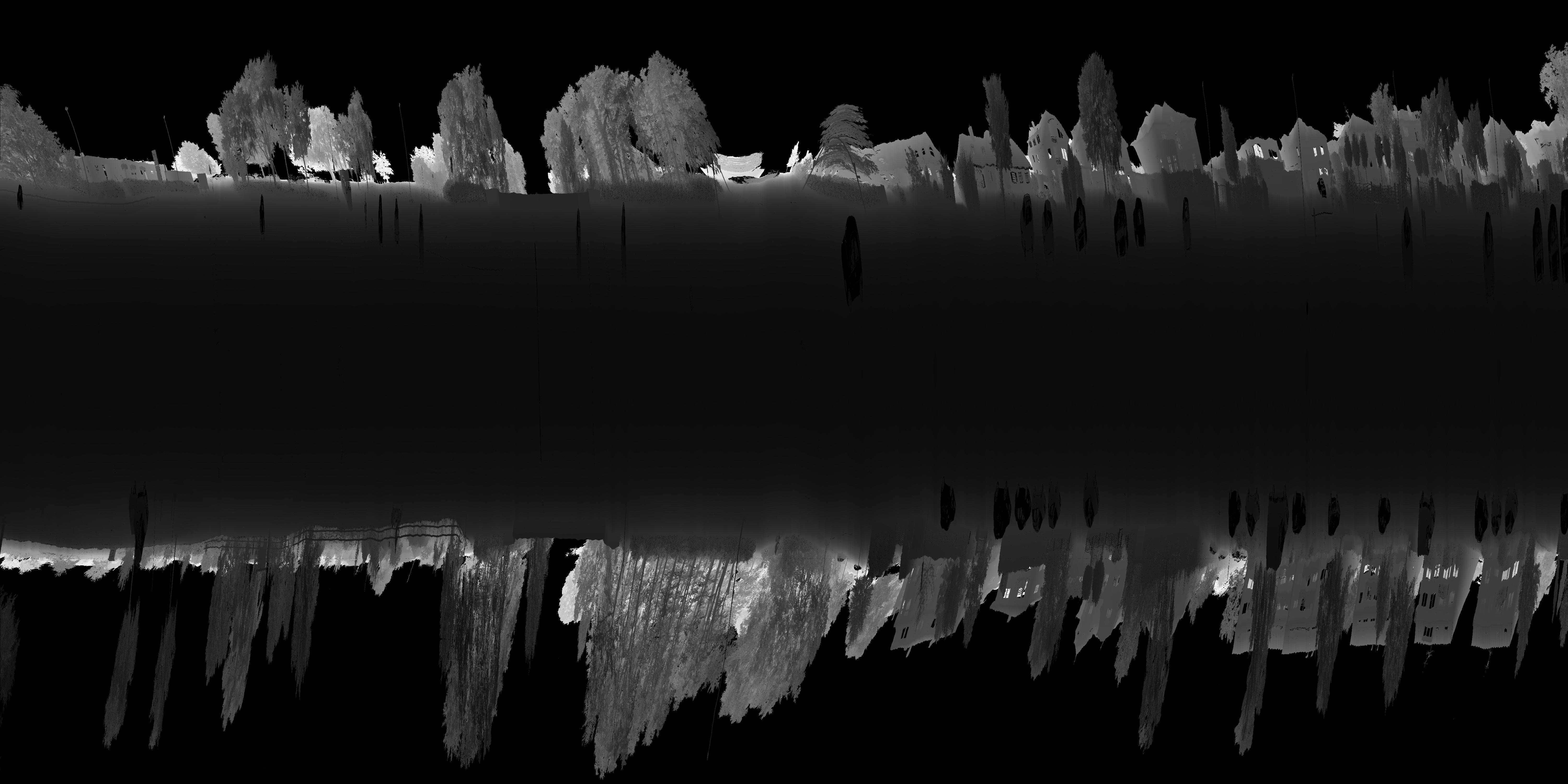}
    \\ (c) scan strip image of distances to the scan head\\
    \end{minipage}%
    }%
    \subfigure{
    \begin{minipage}[t]{.48\linewidth}
    \centering
    \includegraphics[width=\textwidth]{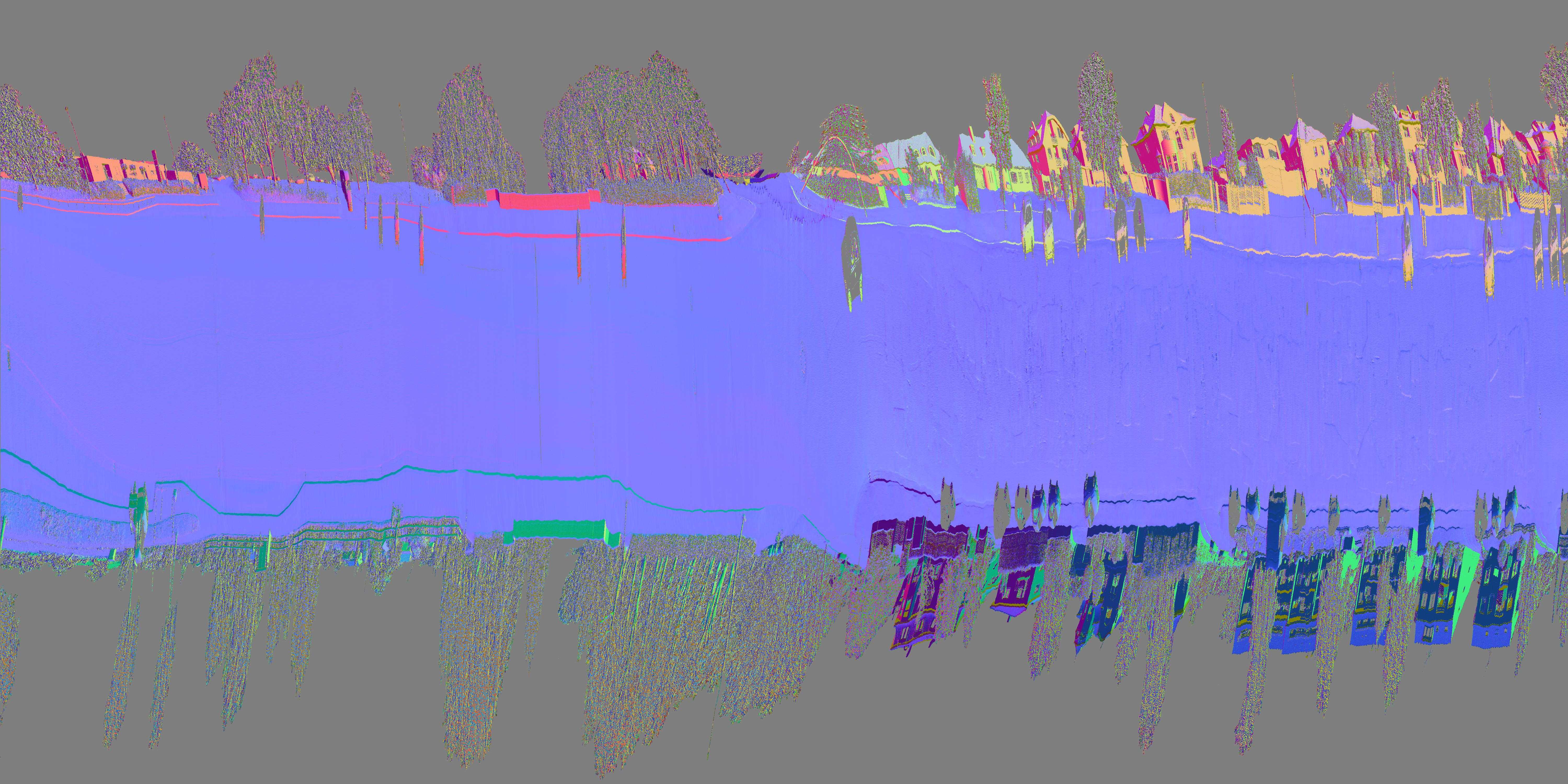}\\ (d) scan strip image of normal vectors\\
    \end{minipage}%
    }%
    \vspace{1mm}
    \subfigure{
    \begin{minipage}[t]{.48\linewidth}
    \centering
    \includegraphics[width=\textwidth]{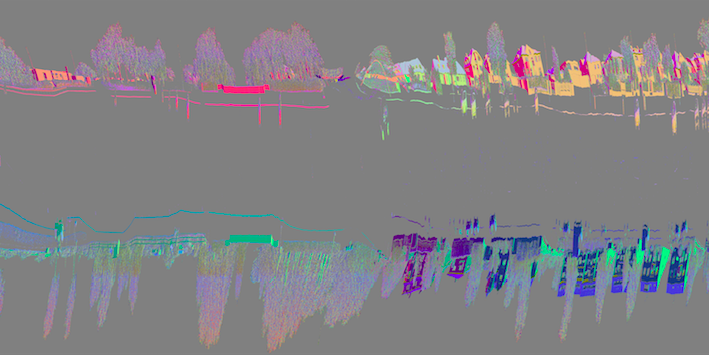}\\ (e) removal of ground points based on normal vectors\\
    \end{minipage}%
    }%
    \subfigure{
    \begin{minipage}[t]{.48\linewidth}
    \centering
    \includegraphics[width=\textwidth]{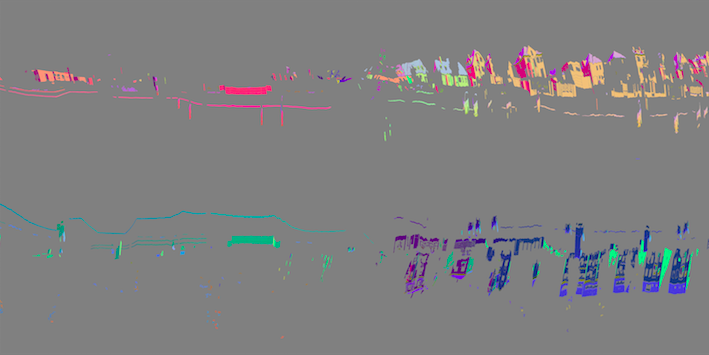} (f) data input for region growing\\
    \end{minipage}%
    }%
    \caption{Visualization of one scan strip at different pre-processing steps.}
    \label{fig:visulization_preprocess_data}
\end{figure*}


The overview of the entire proposed workflow is visualized in Figure \ref{fig:process_overview}, where the main components and their corresponding section numbers are presented. It consists of five major steps.
Building facades are extracted first from adjusted 3D point cloud
and projected to 2D facade images (Section \ref{sec:facade_extraction}).
A rule-based windows and doors detection method (Section \ref{sec:rule_based}) is applied to identify proper examples which can be used for training a deep convolutional neural network 
object detection model (Section \ref{sec:semi_supervised}). 
Post-process\-ing is then introduced, where false detections due to occlusions can be removed based on the reconstruction of laser rays (Section \ref{sec:post_processing}).
Lastly, the detected windows and doors are projected on a 2D map and represented with line segments.
This information is combined with the flood simulation results in order to determine a building flood risk map (Section \ref{sec:indentification}).
All these steps are detailed in the following sections separately.

\subsection{Facade extraction}
\label{sec:facade_extraction}

Due to the complex urban road network, it is inevitable that mobile mapping LiDAR may pass over the same locations several times during the measurement process.
Mainly due to errors in GNSS, the raw measurements of mobile mapping LiDAR may show substantial offsets, in the order of several decimeters, between multiple measurement epochs.
Therefore, the adjustment of scans from different epochs is necessary, especially for large-scale mobile mapping LiDAR measurements. This was achieved by the approach proposed in \cite{brenner2016scalable}, which uses a large-scale least squares adjustment to estimate trajectory corrections based on the 3D points observed during all epochs.

\begin{figure}
    \centering
    \includegraphics[clip, trim=100 150 490 140, width=.32\textwidth]{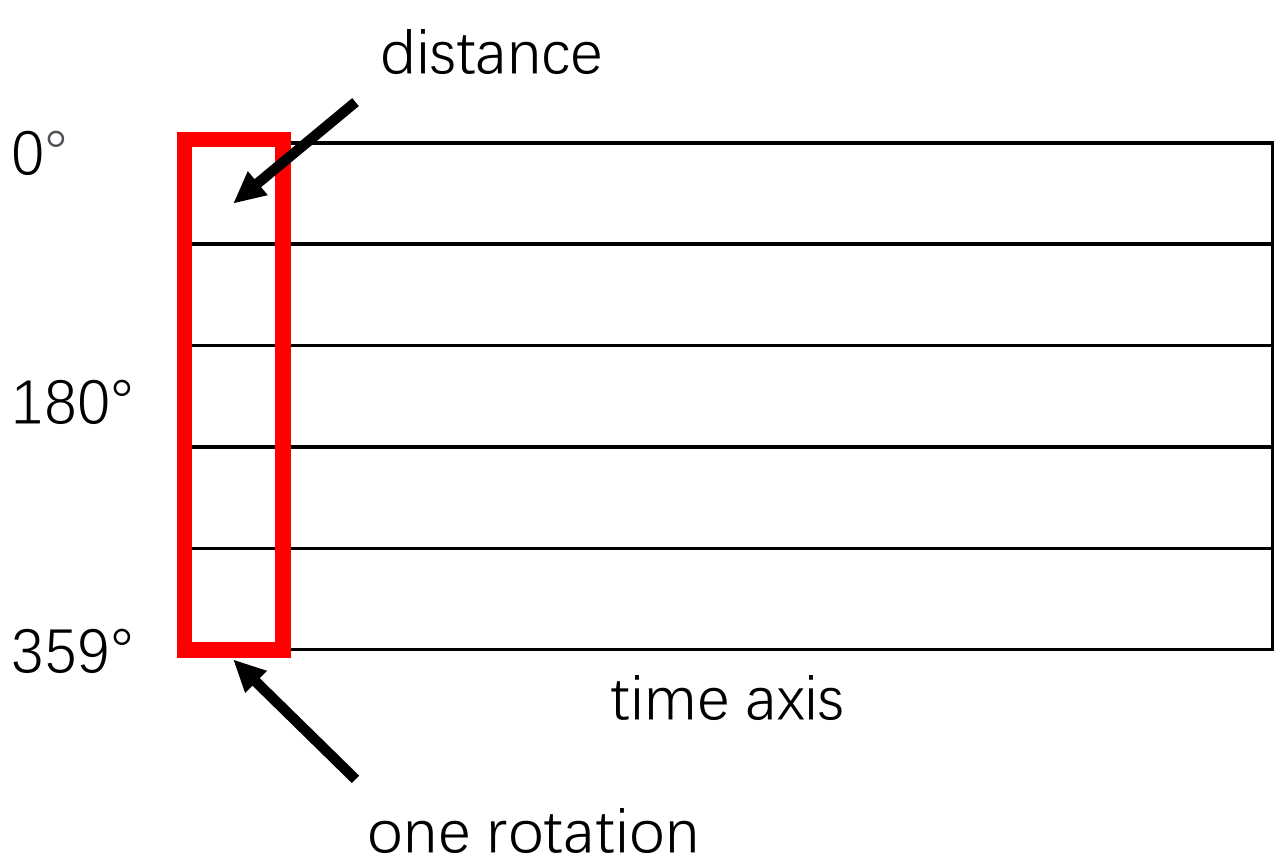}\\
    \caption{An illustration of a scan strip image.}
    \label{fig:scanstrip_image}
\end{figure}

\begin{figure}
  \centering
  \subfigure{
  \begin{minipage}[t]{.48\linewidth}
    \centering
    \includegraphics[clip, trim=0 0 30 0, width=\textwidth]{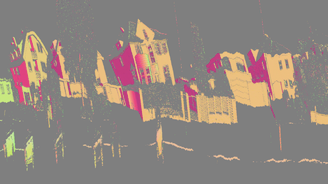}\\ (a)\\
  \end{minipage}%
  }%
  \subfigure{
  \begin{minipage}[t]{.48\linewidth}
  \centering
  \includegraphics[clip, trim=0 0 30 0, width=\textwidth]{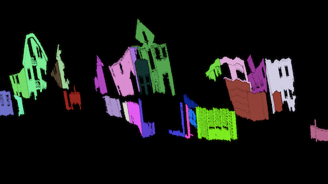}\\ (b)\\
  \end{minipage}%
  }%
  \\
  \subfigure{
  \begin{minipage}[t]{.48\linewidth}
  \centering
  \includegraphics[clip, trim=0 0 30 0, width=\textwidth]{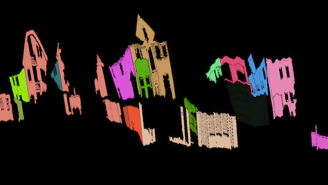}\\ (c)\\
  \end{minipage}%
  }%
  \subfigure{
  \begin{minipage}[t]{.48\linewidth}
  \centering
  \includegraphics[clip, trim=0 0 30 0, width=\textwidth]{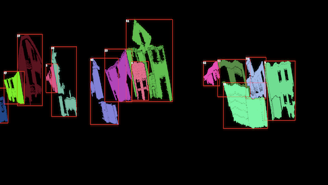}\\ (d)\\
  \end{minipage}%
  }%
  \caption{Process for the extraction of facade instances. (a)~is the normal vectors with $n_z^2 < \theta_n = 0.1$, (b)~is the segmentations based on region growing using both distance and normal vector information. Each segment is rendered by random color. (c)~shows the result of merging of similar and adjacent segments. (d)~shows the final extracted facades after removing fences, low walls, etc.}
  \label{fig:4_fe_facade_instance}
\end{figure}

We base our work on so-called {\em scan strips}, which are rectangular arrays, or images, containing the successive `raw' measurement values in the order in which they were obtained (Figure \ref{fig:scanstrip_image}). Each column represents one full (360 degree) rotation of the LiDAR scanner mirror, and successive full rotations are put into successive columns. Due to the forward movement of the van during data capture, this can also be regarded as a cylindrical projection of the measured 3D points. Most of the time, spatial proximity will be preserved, i.e., cells, or `pixels', which are close in the scan strip will correspond to 3D points which are close in object space, and vice versa. Exceptions occur when there are depth jumps in the scene, or strong heading changes during data capture, e.g. during turns at intersections.
As an example, for the point cloud data presented in Figure \ref{fig:visulization_preprocess_data}(a), the reflectance values and distances to the scan head can be placed in the pixels of this image, as shown in Figures \ref{fig:visulization_preprocess_data}(b) and \ref{fig:visulization_preprocess_data}(c), respectively. In addition, the local surface normal vectors are computed, as visualized in Figure
\ref{fig:visulization_preprocess_data}(d). The normal vector values in XYZ direction are normalized from the range [-1, 1] to the range [0, 1] and assigned to the RGB channels of the image for visualization.

Building facades are further extracted based on the scan strip images of the normal vectors. The facades' normal vectors are large in the horizontal directions, while the normal vectors of the ground are large in the vertical direction. Therefore, a threshold $\theta_n$ is applied on
$n_{z}^2$ to select potential facade points, where $n_z$ is the component of the normal vector in $z$ direction (effectively imposing a constraint on the angle of the normal vector with the `up' vector). The pixels with $n_{z}^2$ below the threshold $\theta_n = 0.1$ are retained as presented in Figure \ref{fig:visulization_preprocess_data}(e). 

With a region growing, the adjacent pixels are grouped to segments based on the cosine similarity of normal vectors with a threshold of 0.95.
In addition, the distances to the scanner head, encoded in each pixel are also considered in the region growing, which can separate adjacent segments, e.g., one from a fence in the foreground and another from the facade in the background.
In order to identify facades, the segments with very few pixels are often outliers, e.g., the points of vegetation. Only the segments with at least 100 pixels are preserved as visualized in Figure \ref{fig:visulization_preprocess_data}(f).


To avoid over-segmentation, the segments created by region growth are re-examined and nearby segments (i.e., not directly connected) with similar averaged normal vector values and distance values are merged, as shown in Figures \ref{fig:4_fe_facade_instance}(b) and \ref{fig:4_fe_facade_instance}(c). Finally, isolated segments containing fences or low walls are removed using a height threshold of 4 meters above the ground. All facades are detected individually, as shown in Figure \ref{fig:4_fe_facade_instance}(d).
The point cloud of a single facade can be computed based on the detected segments.

Each detected segment constitutes a facade, where a representative plane is determined with RANSAC (RANdom SAmpling Consensus).
The projection of the plane to the ground also provides a 2D line segment for each facade, which can be inserted into a map.

%

\subsection{Rule-based window and door detection method using scan lines}
\label{sec:rule_based}

\begin{figure}
    \centering
    \includegraphics[width=.4\textwidth]{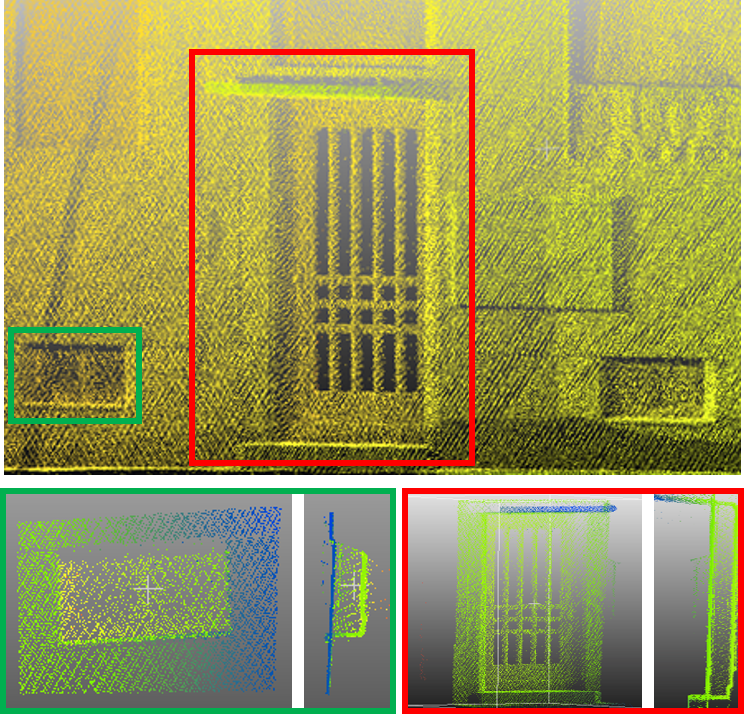}\\
    \caption{Example of hollow and regressed patterns of a window (left) and door (right) on a facade (front and side view).}
    \label{fig:5_rbm_hsp}
\end{figure}

\begin{figure*}
    \centering
    \includegraphics[clip, trim=15 10 10 0, width=\textwidth]{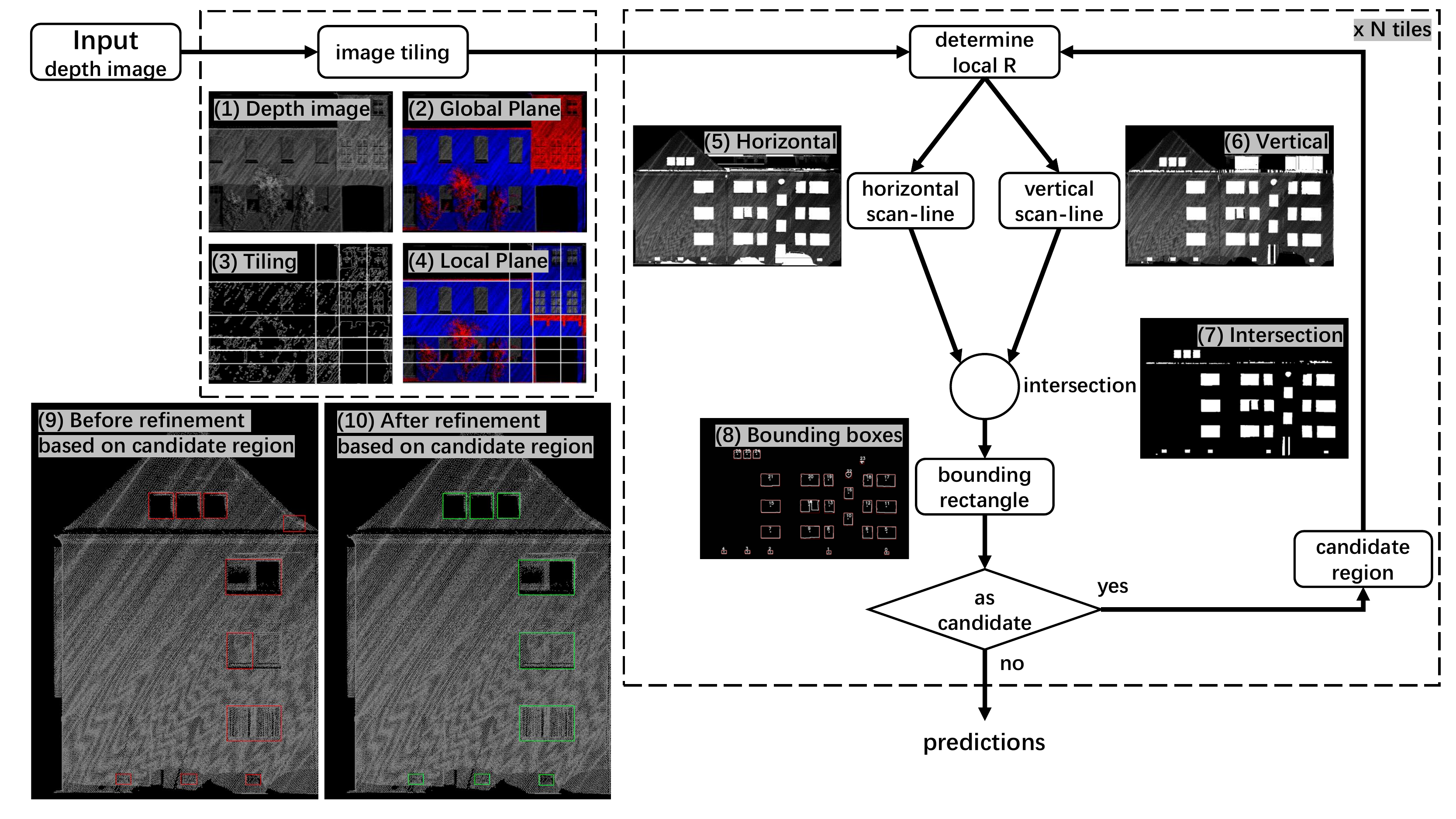}\\
    \caption{An overview of the rule-based method using scan lines. (1)~input depth image. (2)~if the detection is based on a global plane, windows on the building's protruding part would be ignored. The blue areas are regions close to the estimated plane while the red areas are protruding. The rest of the area is recessed or without measurements. (3)~images are divided into tiles based on a Canny edge detection. (4)~reference depth values are calculated for each tile, which allows the detection on the protruding part. (5)~images segmented using horizontal scan lines. (6)~images segmented using vertical scan lines. (7)~segments extracted from horizontal and vertical scan line images are intersected. (8)~bounding boxes are generated after removing very small and isolated detections. (9)~detections are refined locally. Results before refinement have false detections and are incomplete. (10)~false detections are corrected after the refinement step.
    }
    \label{fig:rule_based}
\end{figure*}

\begin{figure}
    \centering
    \includegraphics[width=.48\textwidth]{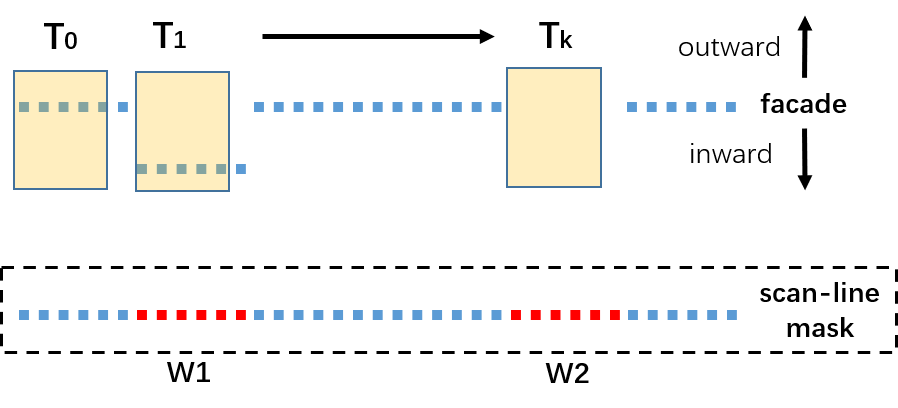}\\
    \caption{Illustration of a row of depth images.}
    \label{fig:scan_line}
\end{figure}

As observed in the point cloud data, doors and windows are mostly regressed into the facades or are non-reflective areas forming holes (see Figure \ref{fig:5_rbm_hsp}). Thus, a rule-based method is developed mainly based on this assumption. 
The entire process is visualized step by step in Figure \ref{fig:rule_based}. 
The inputs of this rule-based detection method are the depth images, as shown in Figure \ref{fig:rule_based}(1).
Depth images are calculated based on the estimated vertical facade plane from Section \ref{sec:facade_extraction} with a cell size of 2~cm. Outward-facing points are assigned with positive values, inward-facing ones with negative values, as presented with red and white colors in Figure \ref{fig:rule_based}(2). If only this globally estimated plane were used to find the facade openings,
windows on the building's protruding parts would be ignored. Therefore, the entire image needs to be divided into tiles, and recessed areas are searched locally to avoid this issue. A Canny edge detector \citep{canny1986computational} is applied to the depth images to find proper splitting locations. 
For each row and column, we count the number of pixels of the binary edge image. The top twenty rows and columns are selected in both horizontal and vertical directions as the location of cutting lines. However, very close cut lines should be avoided. Through our observation, a minimum vertical spacing of 1.6~m and a horizontal spacing of 1.0~m should be maintained to avoid overly dense cut lines. An example of the generated cutting lines to generate image tiles is visualized in Figure \ref{fig:rule_based}(3). Subsequently, planes are estimated, again using RAN\-SAC, for individual tiles, which enable the detection of locally recessed areas, as shown in Figure \ref{fig:rule_based}(4).

Scan line grouping is a classical method for the segmentation of range images, dating back to the work of \cite{jiang1994fast}. 
The basic idea is to apply horizontal scan lines with regular spacing and break a single scan-line into line segments. After scanning the entire facade, the adjacent line segments with similar characteristics are combined, similar to the region growing only in terms of line segments.
Windows and doors are mostly rectangular in shape. Therefore applying scan lines on both horizontal and vertical directions can provide good candidates of recessed areas on the facades.
Since windows and doors are mostly no smaller than 10~cm and mostly recessed into the facades, a sliding window with a five pixel size is applied on each row or column of the depth image. 
Once five successive pixels are below the facade plane or missing, they are marked as objects as illustrated in Figure \ref{fig:scan_line}. 

Afterwards, detected object lines in the horizontal and vertical directions are grouped as presented in Figure \ref{fig:rule_based}(5) and Figure \ref{fig:rule_based}(6). It can be observed that the missing parts on the facades due to occlusions are marked on the images generated by horizontal scan lines, and the windows on the roofs without measurements on the bottom bounds are marked on the images generated by the vertical scan lines. Therefore, in order to achieve better precision, the intersection of both masks is calculated as shown in Figure \ref{fig:rule_based}(7) and used for generating detection bounding boxes, as shown in Figure \ref{fig:rule_based}(8). 
Standard image processing is used to identify objects and the corresponding bounding boxes (OpenCV method \textit{findContours} \citep{suzuki1985topological}). Very small or elongated detections are filtered out using thresholds.

Still, the results of this rule-based method are not yet satisfying due to varying structures on the facades. Therefore, the detections from the previous step are only used as candidates. The same procedure using scan lines is performed again based on each candidate, with an enlarged scope of three times the maximum side length of the candidate box. The benefits of this refinement are presented in Figures \ref{fig:rule_based}(9) and \ref{fig:rule_based}(10). Many of the false detections and incomplete detections are corrected in this step.



\begin{figure*}
  \centering
  \includegraphics[height=0.23\textwidth]{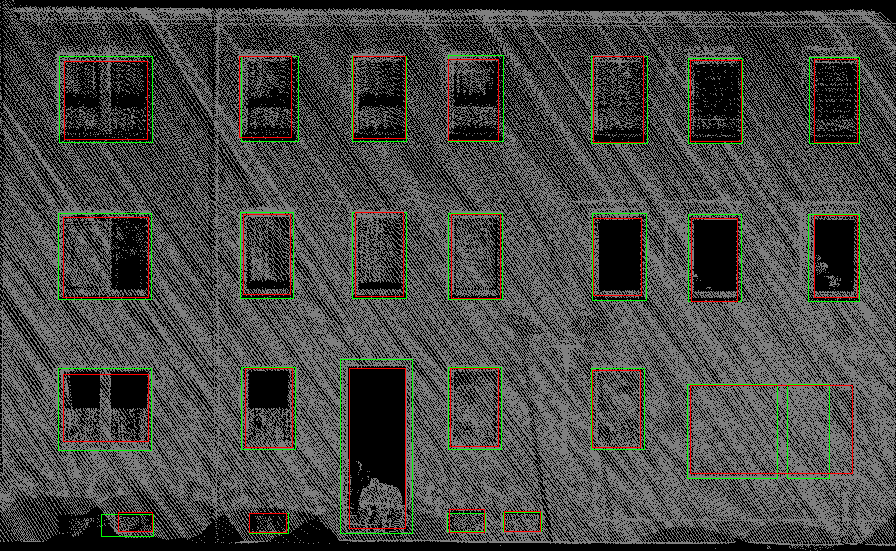}
  \includegraphics[height=0.23\textwidth]{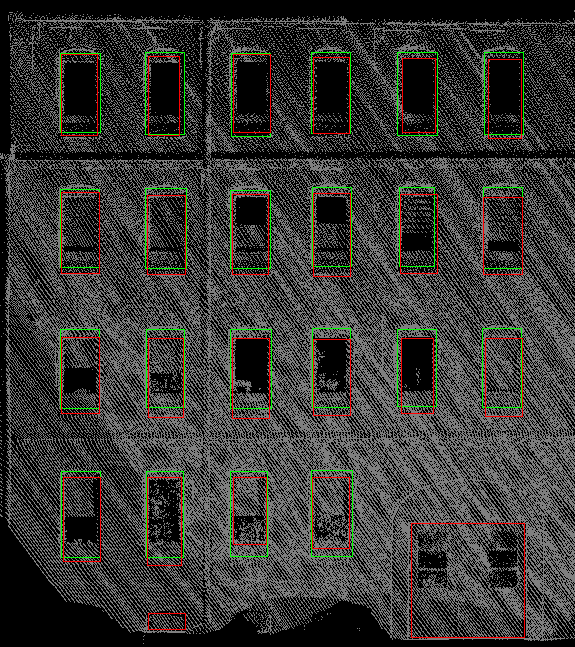}
  \includegraphics[height=0.23\textwidth]{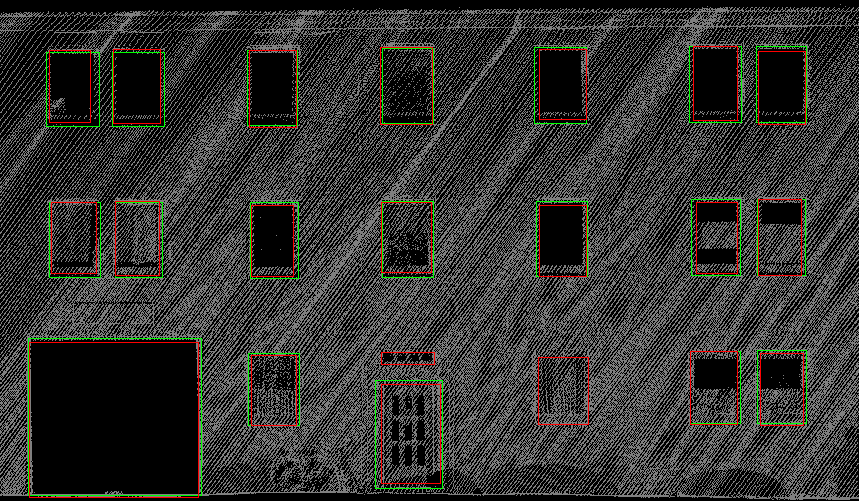}
  \caption{Comparison of pseudo-labeled (green) and manually annotated (red) examples.}
  \label{fig:comp_peudo}
\end{figure*}

\begin{figure}
    \centering
    \includegraphics[clip, trim=450 0 0 0, width=.48\textwidth]{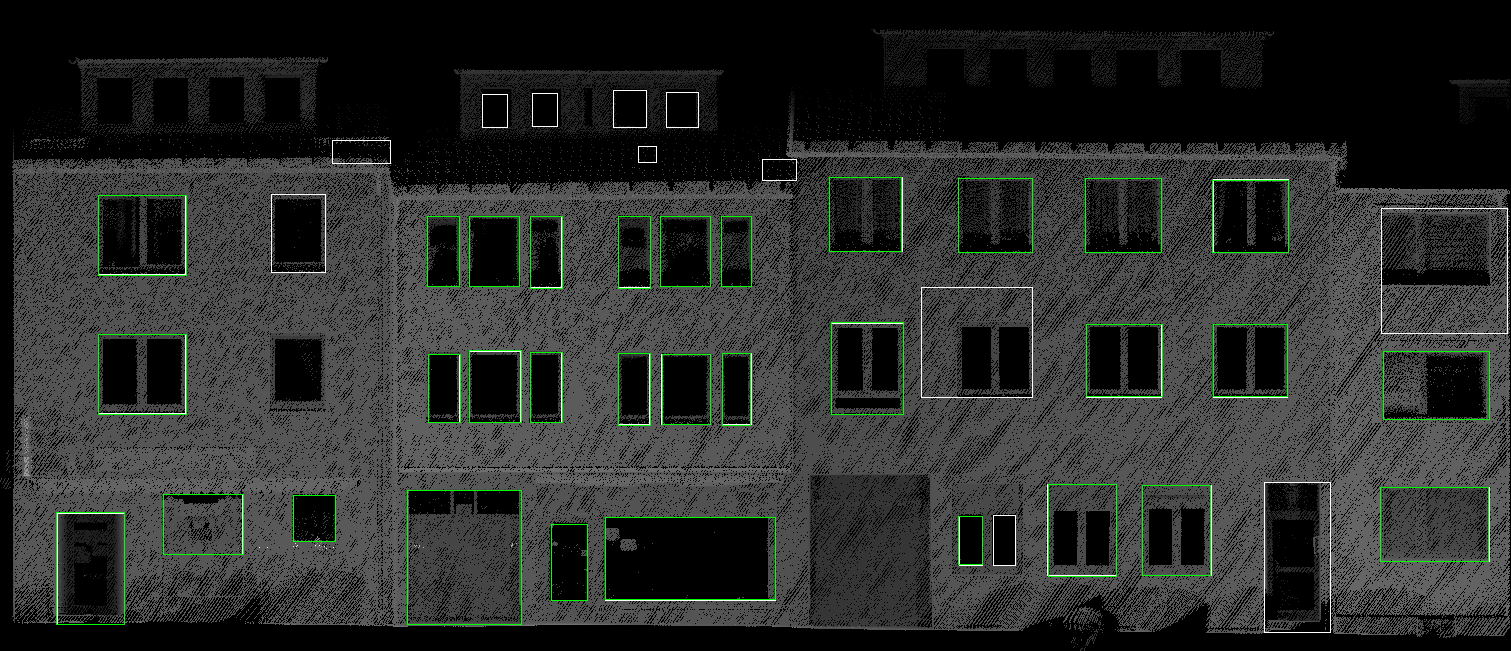}\\
    \caption{The white boxes are all predictions of the rule-based method, while the green boxes are the predictions satisfying the condition $IoU>0.8$, which are later used as training samples.}
    \label{fig:filter_trainset}
\end{figure}

Even though the detection method using scan lines achie\-ved good performance in most cases and reached a precision of 65.2\% in the later experiments, this rule-based method requires several pre-defined parameters to describe the facade openings. Therefore, this method can hardly cover all situations in complex urban scenarios.
However, they can serve as input for training a deep learning model.

Before using them as training examples, however, potentially incorrect detections should be first filtered out.
Our rule-based method applied a refinement step after the object detections on individual tiles.
According to our observations, most of the detections with only a small change before and after this refinement step are mostly of better quality. Therefore, a threshold is applied. Only the boxes with an IoU (Intersection over Union) before and after refinement greater than 0.8 are preserved as illustrated in Figure \ref{fig:filter_trainset}. 
IoU is a measure to describe the overlap of two bounding boxes, which is defined as the ratio between the area of their intersection and the area of their union.
In this way, pseudo labels with relatively better quality are preserved.
Figure \ref{fig:comp_peudo} shows a comparison between the pseudo-labeled (green) and manually annotated (red) examples.

\subsection{Deep learning models with and without additional supervision}
\label{sec:semi_supervised}

\begin{figure}
    \centering
    \includegraphics[clip, trim=60 180 200 60, width=.48\textwidth]{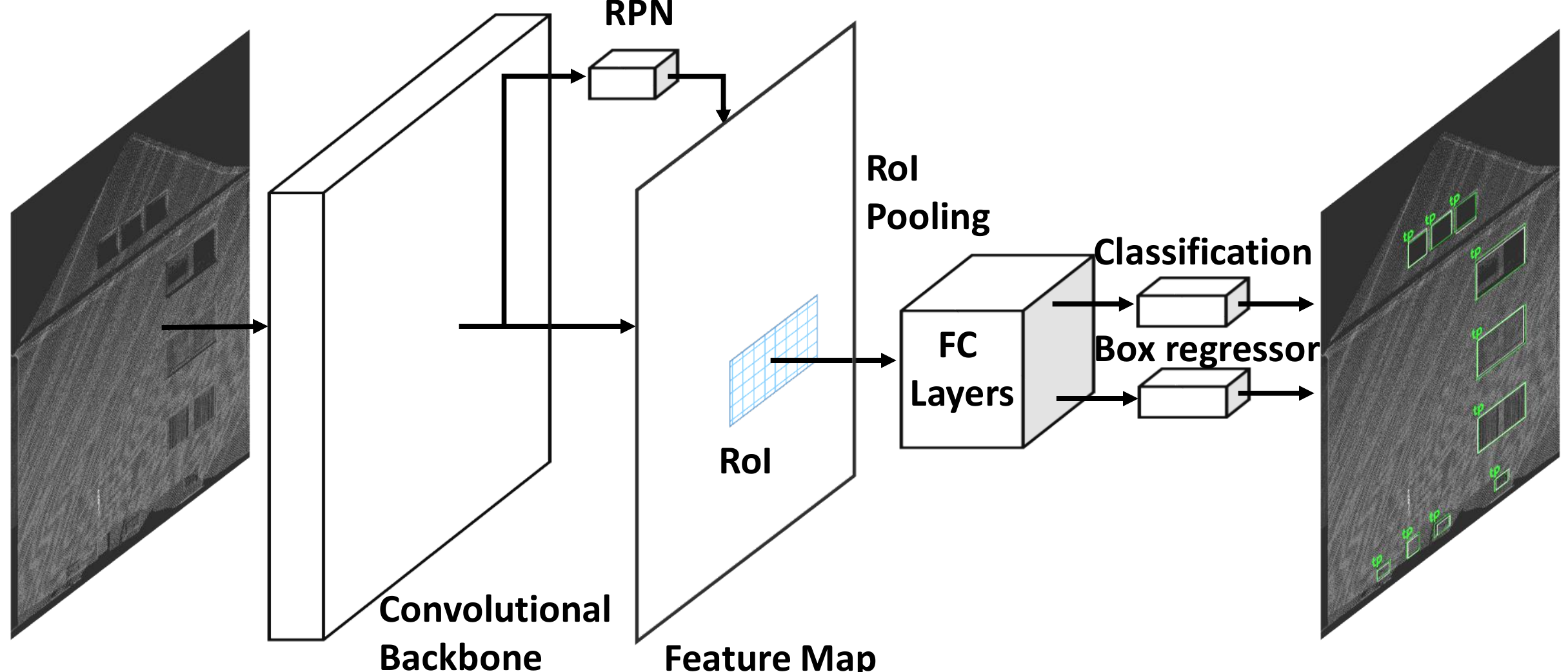}\\
    \caption{Network architecture of Faster R-CNN.}
    \label{fig:faster_rcnn}
\end{figure}

Faster-RCNN \citep{ren2015faster} is one of the well-known two-stage deep convolutional neural network architectures for object detection.
The whole network can be roughly divided into five modules: feature extraction layers (i.e., backbone network), Region Proposal Network (RPN), ROI-Pool\-ing, classifier branch, and box regressor branch, as illustrated in Figure \ref{fig:faster_rcnn}.
The backbone used in this research is ResNet50 using Feature Pyramid Network (FPN). Unlike the common ResNet50, which produces one single feature map, FPN applies an encoder-decoder-like network structure, which outputs image deep features for region proposal at multiple scales.
FPN is a structure that proved to be beneficial for detecting small objects \citep{lin2017feature}. RPN predicts a set of regions of interest (ROIs) as rectangular boxes. ROI-pooling is used to map the deep features of ROIs in arbitrary shapes to fixed-size feature maps. With the output feature maps, the objects are classified, and the object bounding boxes are refined.

The input data of this model should include significant features about the windows and doors. Therefore, three information sources are considered in this work: the depth with respect to the estimated global plane, the point density, and the reflectance.
The density is calculated by summarizing the number of points in the 2D projected image and stored in point density images. As there are often complex structures adjacent to doors and windows, usually the density of points in the 2D-projection in these areas is higher than the rest of the facade areas. This image is expected to provide information about the edges of the facade openings.
Lastly, reflectance incorporates the physical properties of different materials, i.e., the facades have a significantly different value range than the other objects, e.g., curtains behind the windows or the metal covers in front of the basement windows.

The neural network is then trained based on the composite image, which includes these three channels (see the example in Figure \ref{fig:process_overview}). 
Since the input channels have different value ranges, normalization is applied for each channel individually using
    $x' = (x-\mu)/\delta$
where $\mu$ is the mean and $\delta$ is the standard deviation of the individual channels over the entire dataset.

We used the pytorch implementation of Faster-RCNN\footnote{Torchvision Models. \url{https://pytorch.org/vision/stable/models.html} (Accessed on 08.03.2021)}.
%
Regarding parameter settings, smaller anchor scales (i.e., 2, 4, 8, 12) are used instead of the default scales of the COCO dataset (i.e., 32, 64, 128, 256, 512).
We further checked the predictions to confirm that these anchor scales could cover all of our targets essentially. Anchor aspect ratios have been extended to 0.33, 0.5, 1.0, 2.0, 3.0. 
We considered all possible combinations of anchor scales and anchor aspect ratios, i.e., 20 anchor boxes are generated for each position on the feature map.

\subsection{Post-processing using laser ray tracing}
\label{sec:post_processing}

\begin{figure}[h]
    \centering
    \includegraphics[width=.48\textwidth]{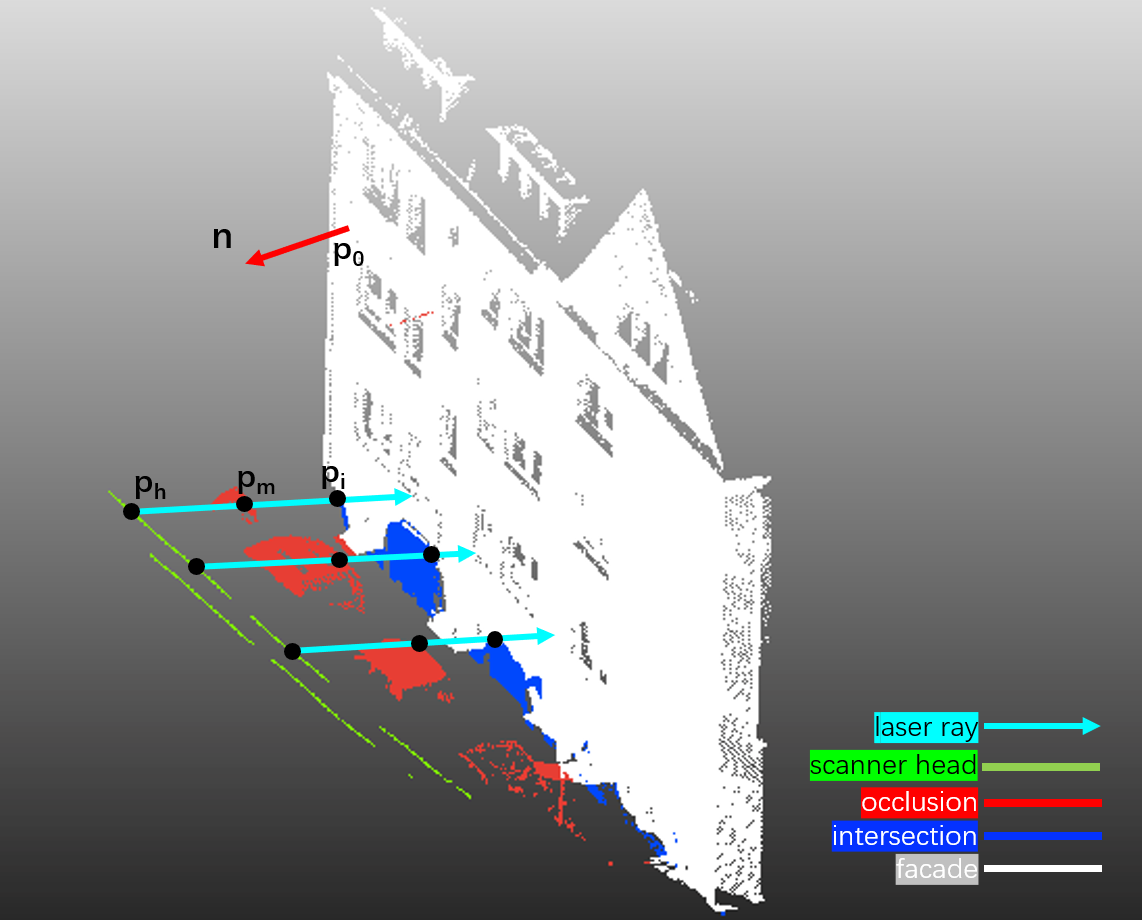}\\
    \caption{The intersection points of laser rays and a facade.}
    \label{fig:post_processing}
\end{figure}

Since there are still several false detections observed, especially in places where the facades are occluded by objects in front of them, we perform a hole filling based on laser ray tracing.
As illustrated in Figure \ref{fig:post_processing}, the red points are objects in front of the facade, leading to occlusions. Since for each measured 3D point, the corresponding position of the scanner head (in green), at the time the laser pulse was emitted, is known, the exact laser ray can be reconstructed. For occluding objects, which are identified during the pre-processing step, the ray can be extended until it intersects the facade plane (filled facade parts, in blue).
Using this, an occlusion mask can be generated for each facade image, which is later used to reject improper detections due to occlusions.

\subsection{Identification of building's flood risk}
\label{sec:indentification}

With the detections of deep learning models, line segments of doors and windows are marked on a 2D map with the attributes of the height of their bottom edge. 
In order to obtain a height relative to the ground, it is necessary to extract the ground from the point cloud data or find other sources of information for areas where no mobile mapping LiDAR data is available on the ground. Measurement points on the ground adjacent to buildings are sometimes missing due to occlusions, e.g., grass or fences.
There are also buildings that are obscured by the ones in the front, where only the higher facade parts can be scanned.
Therefore, many facades can not be associated with a proper ground height using mobile mapping LiDAR acquisitions alone.

Our previous work \citep{feng2018enhancing} allowed us to generate high-resolution terrain information and reasonably integrate ground measurements from mobile mapping LiDAR and the airborne laser scanning derived DTM from the mapping agency. It produces a terrain model which is complete and smooth. This allows a ground reference height to be found for each detected target.

The relative heights of the facade openings above terrain are calculated and compared with the forecasting results from a 2D hydraulic simulation of surface water runoff. The flood forecasting model used in this research is a bi-directionally coupled pipe-overland flow simulator from BPI Hannover\footnote{BPI Hannover - Engineering Office for Urban Water Management. \url{https://bpi-hannover.de/} (Accessed on 08.03.2021)}. Flow in the urban pipe network is calculated with an in-house code, similar to the widely used model storm water management model \citep{gironas2010new}, where pipe flow is conceptualized as open-channel flow using the de Saint Venant equation system \citep[e.g.][]{Yen1973}. Pipe flow is solved using the Finite-Volume method. Overland flow is calculated with the multiphysics simulator OpenGeoSys \citep{Kolditz2012,delfs2013coupled,jankowski2021overland} using the diffusive wave equation and the centered Galerkin Finite-Element-Method. The two simulators are bidirectionally coupled by timestep-wise calculating exchange fluxes at manholes according to \citet{rubinato2017experimental}. Similar to the coupling described in \cite{peche2017coupled}, exchange flows are timestep-wise added or subtracted from the respective (pipe- and overland flow) continuity equation. Thus, flood simulation results in the present study are based on coupled pipe and overland flow.


Since the calculation of the flood forecasting model is based on a Triangulated Irregular Network (TIN), the output results are simplified as points with spacing roughly around 6-8m. For each detected edge, the nearest water level prediction point is searched to calculate the flood risk.
We consider that the higher the estimated water level is above the detected targets' bottom edge, the greater the risk.
A flood risk index $I$ is therefore defined as
\begin{equation*}
    I = H_{water} - H_{window}
\end{equation*}
where $H_{water}$ is the predicted water level above the ground and $H_{window}$ is the height of the individual windows and doors above the ground.

Since our detections have an uncertainty, only the edges with $I$ above 6~cm are considered to be a flood risk, which corresponds to the average height error of the detection models. With this information, edges are visualized with different building flood risk levels with respect to the index $I$.

\section{Experiments and analysis}
\label{sec:experiments}

In the following, the above-presented process has been applied to a test site \textit{Neustädter Markt}, located in Hildesheim, Germany. The mobile mapping system Riegl VMX-250\footnote{Riegl VMX-250 Data Sheet. \url{http://www.riegl.com/uploads/tx_pxpriegldownloads/10_DataSheet_VMX-250_20-09-2012.pdf} (Accessed on 08.03.2021)} was used. It contains two Riegl VQ-250 laser scanners, which can measure 600,000 points per second. Positions and orientations are provided by GNSS and IMU, which are corrected by the reference data provided by SAPOS\footnote{SAPOS Service (German). \url{https://www.lgln.niedersachsen.de/startseite/geodaten_karten/geodatische_grundlagennetze/sapos/sapos-services-und-bereitstellung-143814.html} (Accessed on 08.03.2021)} in Lower Saxony, Germany. A total of 98 scan strips were obtained.

\subsection{Pre-processing}
\label{sec:data_acquisition}

\begin{figure}
    \centering
    \includegraphics[width=.48\textwidth]{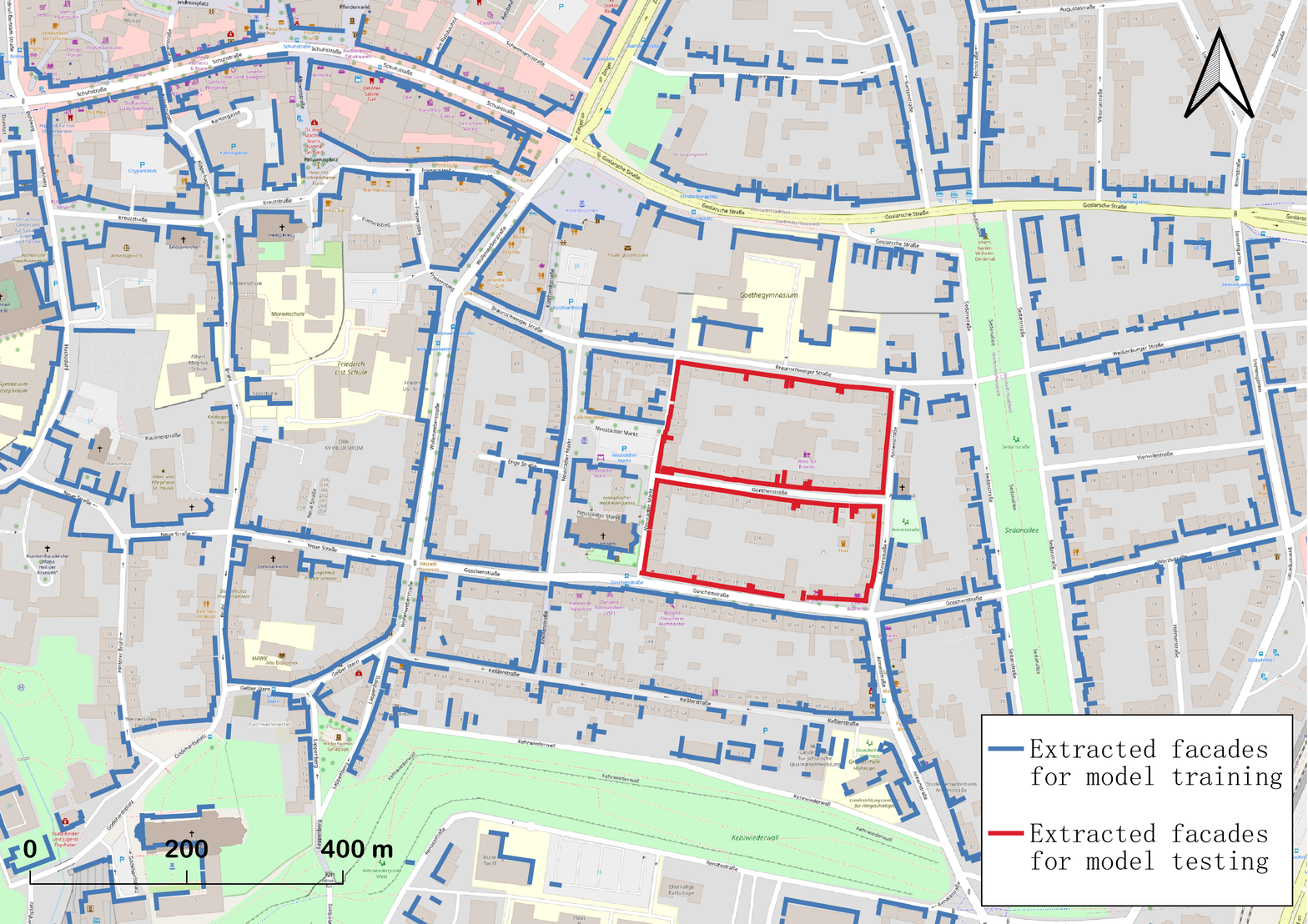}\\
    \caption{Overview of the extracted facades in the study area. The red facades were manually annotated and used for model testing and comparison. The rule-based method was applied for the remaining facades (in blue), which are used for the semi-supervised learning. (Basemap: OpenStreetMap)}
    \label{fig:overview_map}
\end{figure}

The adjustment of scan strips is performed with the process proposed in \cite{brenner2016scalable}.
%
%
For each adjusted scan strip, facades are detected with the steps presented in Section \ref{sec:facade_extraction}. A total of 2299 facades were extracted for the entire study area. The corresponding line segments of the extracted facades are presented on the map in Figure \ref{fig:overview_map}. 

\subsection{Window and door detection}

In order to evaluate the method for the detection of windows and doors, the 2D projected facade images marked in red in Figure \ref{fig:overview_map} were manually annotated and used for testing and model comparison. 
It contains 92 facade images with 1428 annotated objects.
This is an area representative for the whole study area, with a mix of different types of buildings.
The windows and doors on the facade images are manually annotated using bounding boxes. Precision, recall, and \Fone-score are the basic performance metrics for object detection tasks. The detection which has an IoU above 0.5 is regarded as a positive detection. In addition, the bottom edge heights of the detected objects are expected to be precise for our application scenarios. Therefore, a height error is defined by calculating the average height difference of the true-positive examples. It is defined as
\begin{equation*}
E_h = \frac{1}{N}\sum_{i}^{N}{|h^i_g-h^i_p|}
\end{equation*}
where $h^i_g$ is the ground truth height of the $i$-th box, $h^i_p$ is the corresponding predicted height, and $N$ is the total number of true-positive examples.

\subsubsection*{Ablation studies for the rule-based method}

The rule-based method using scan lines is first evaluated with this test set. In the process introduced in Section \ref{sec:rule_based}, many parameters are defined to describe the shape and characteristic of the facade openings. Most of these parameters are chosen empirically and are difficult to be tested comprehensively, because it is possible that a set of parameters that performs well in one region may not perform well in other regions.

The rule-based method applies a refinement step for the facade openings detected on image tiles. Therefore, we conducted an ablation study by comparing the results with and without this refinement step. A qualitative comparison is presented in Figures \ref{fig:rule_based}(9) and \ref{fig:rule_based}(10). A quantitative comparison of the test set is presented in Table \ref{tab:scan_line_ablation}.
It can be seen that the refinement results in a higher precision and \Fone-score. The recall rate is slightly reduced because no new targets can be detected by introducing this refinement step. The average height error of the targets' bottom edge is significantly decreased.

In summary, a precision of 0.645 could be achieved by this model, which can be used to provide training data for the learning-based model.
Despite the best efforts to optimize the model, these rules are difficult to generalize and perform poorly in some scenarios, especially in low point density scenarios.
Curtains, items on windowsills, and window stickers are also common causes for incomplete detections.

\begin{table}[H]
  \centering
    \begin{tabular}{lcccc}
    \toprule
    Ablations  & Prec. & Rec. & \Fone-score & $E_h$ [cm] \\
    \midrule
    w/o refinement & 0.562 & \textbf{0.668} & 0.610 & 6.91  \\
    with refinement & \textbf{0.645}  & 0.654 & \textbf{0.650} & \textbf{6.03 }\\
    \bottomrule
    \end{tabular}%
  \caption{Ablation study for the introduction of the refinement step.}
  \label{tab:scan_line_ablation}%
\end{table}

\begin{figure*}
  \centering
  \subfigure{
    \begin{minipage}[t]{0.235\linewidth}
    \centering
    \includegraphics[clip, trim=410 20 20 20, width=\textwidth]{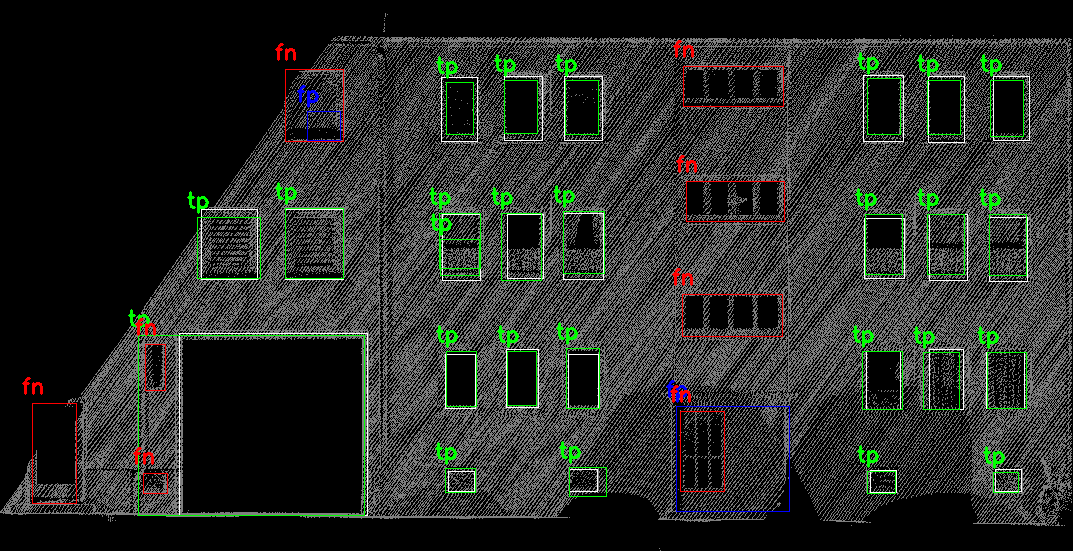}\vspace{0.5mm}\\
    \includegraphics[clip, trim=230 20 340 100, width=\textwidth]{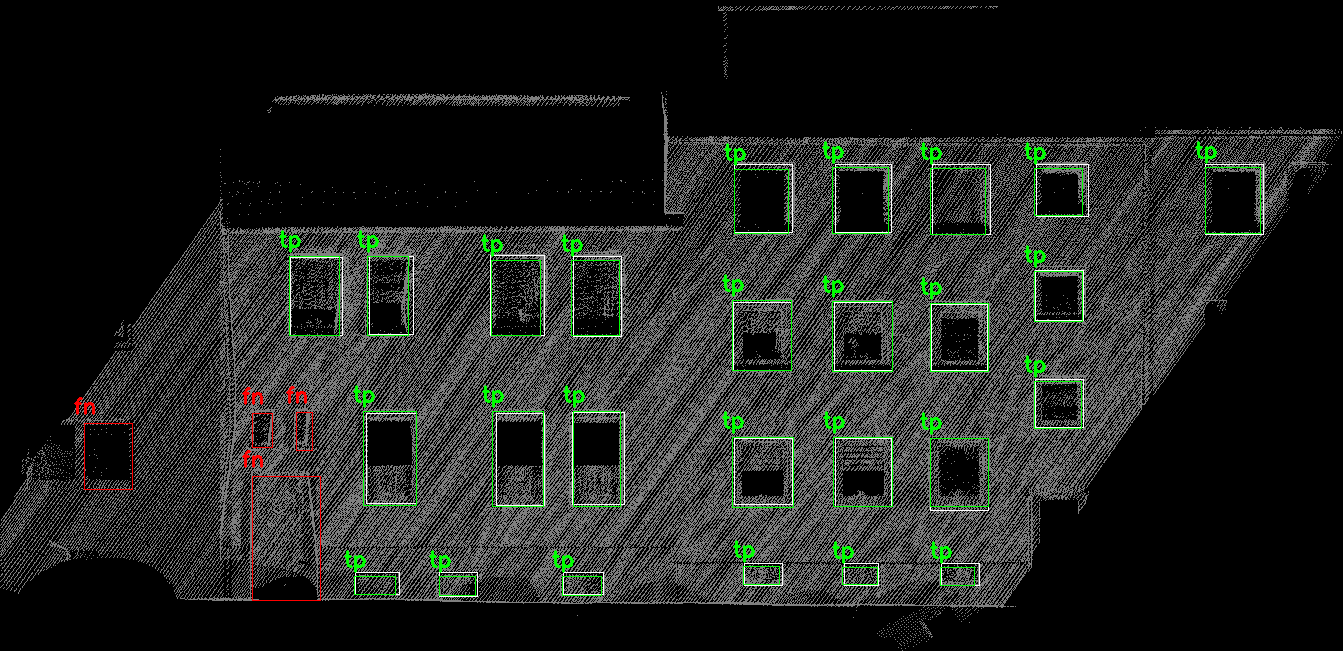}\vspace{0.5mm}\\
    \includegraphics[clip, trim=330 0 240 70,width=\textwidth]{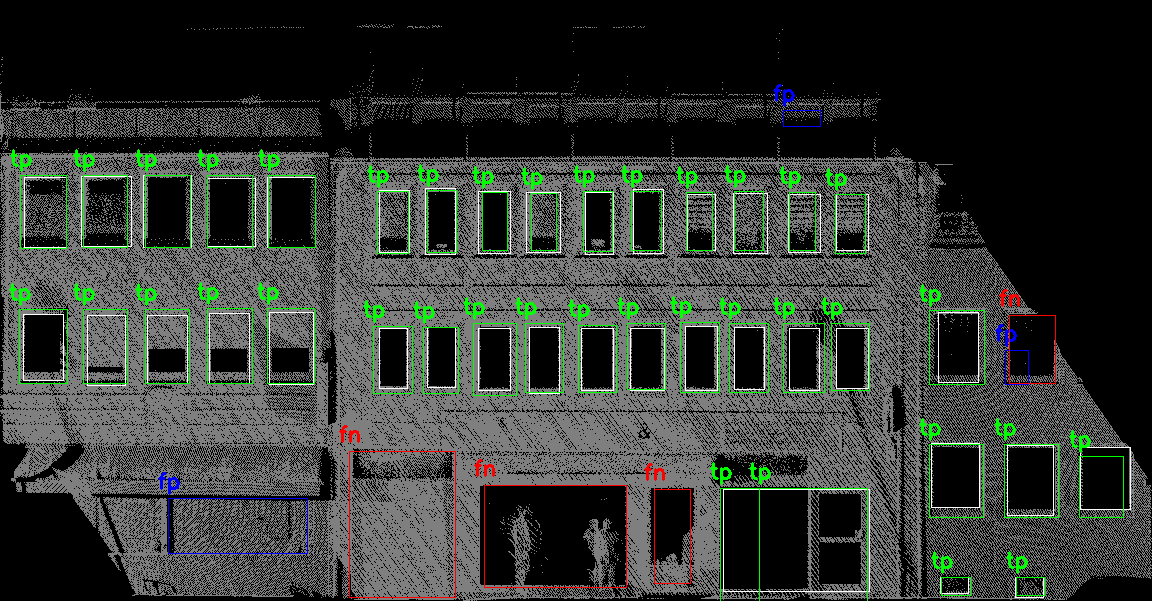}\vspace{0.5mm}\\
    \includegraphics[clip, trim=940 0 0 0,width=\textwidth]{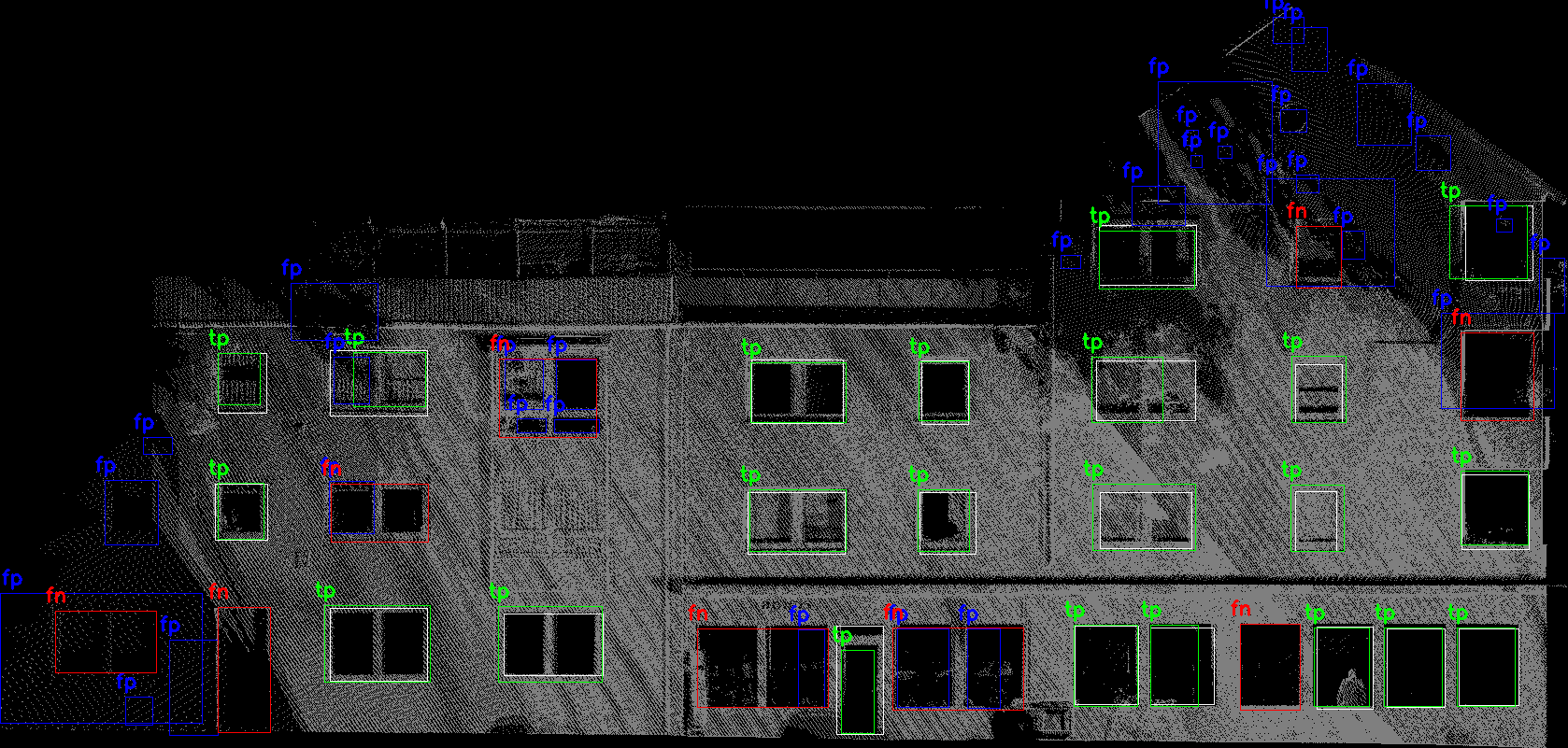}\vspace{0.5mm}\\
    \includegraphics[clip, trim=0 0 0 0,width=\textwidth]{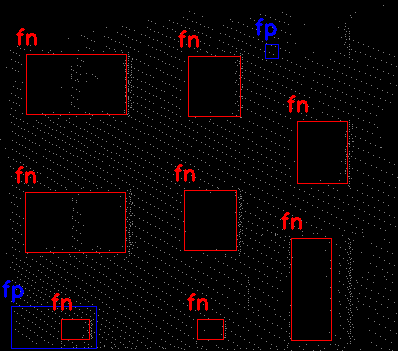}\vspace{0.5mm}\\
    \includegraphics[clip, trim=0 60 100 25,width=\textwidth]{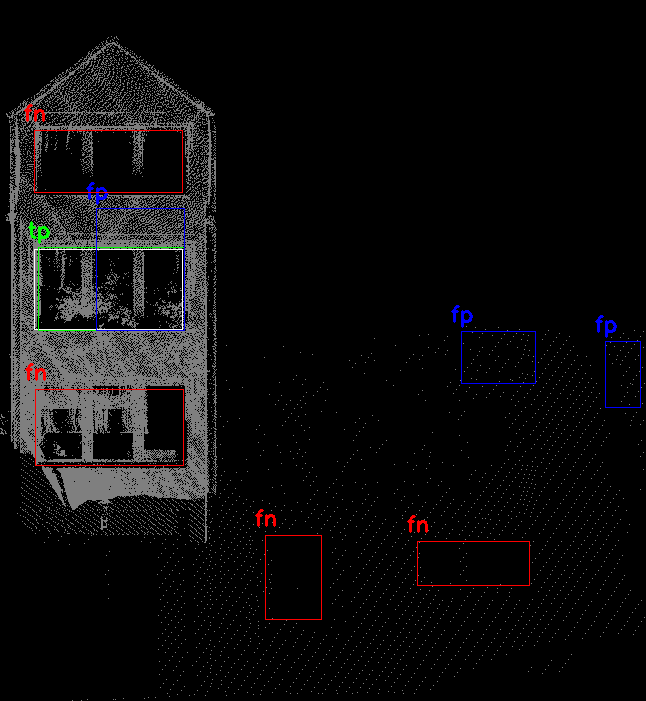}\\
    RULE-BASED\\
    \end{minipage}%
    }%
    \hspace{1mm}
  \subfigure{
    \begin{minipage}[t]{0.235\linewidth}
    \centering
    \includegraphics[clip, trim=410 20 20 20,width=\textwidth]{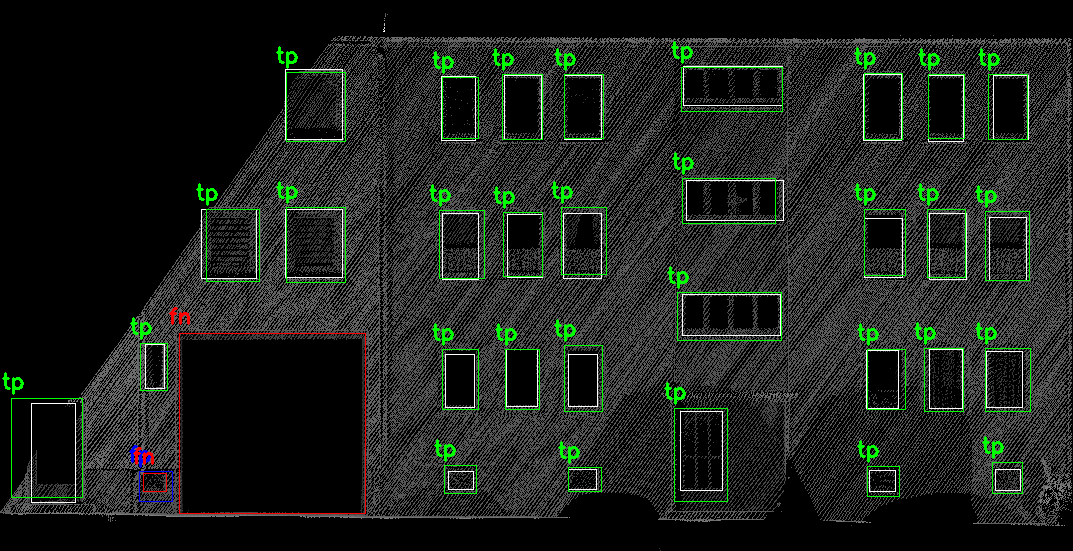}\vspace{0.5mm}\\
    \includegraphics[clip, trim=230 20 340 100,width=\textwidth]{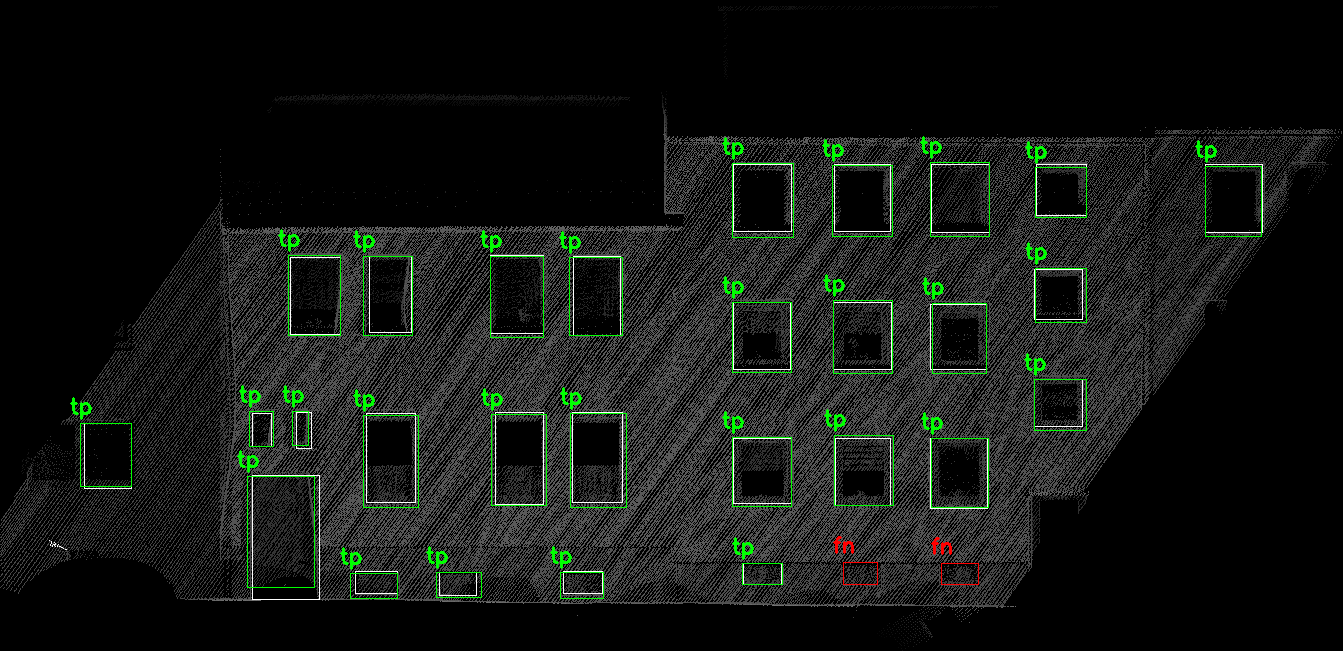}\vspace{0.5mm}\\
    \includegraphics[clip, trim=330 0 240 70,width=\textwidth]{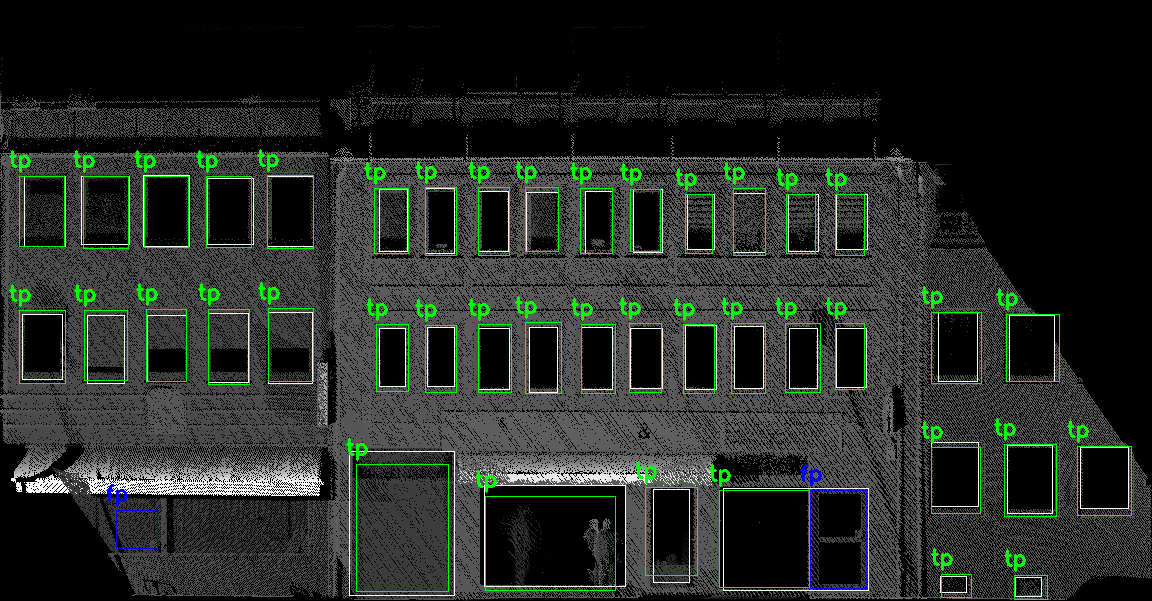}\vspace{0.5mm}\\
    \includegraphics[clip, trim=940 0 0 0,width=\textwidth]{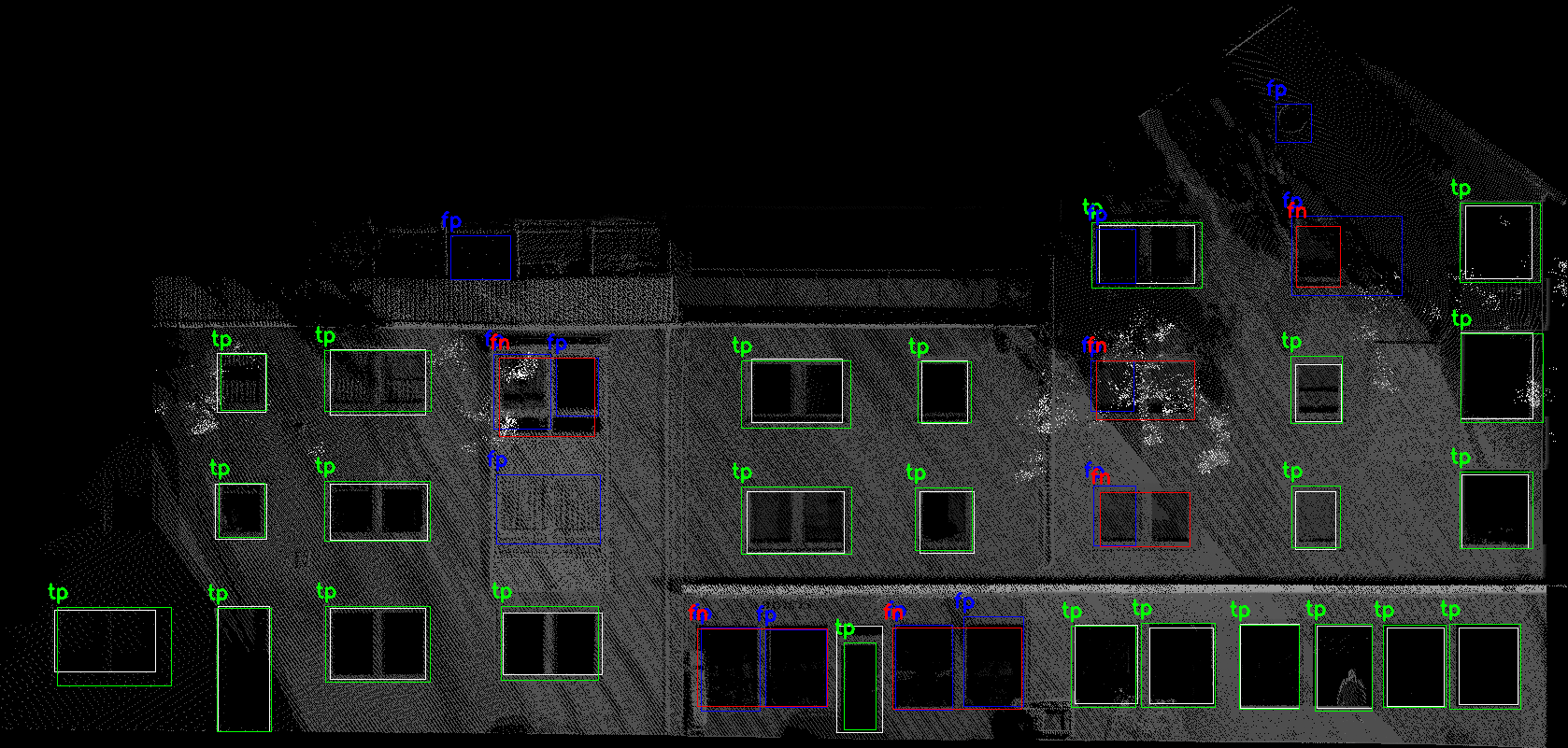}\vspace{0.5mm}\\
    \includegraphics[clip, trim=0 0 0 0,width=\textwidth]{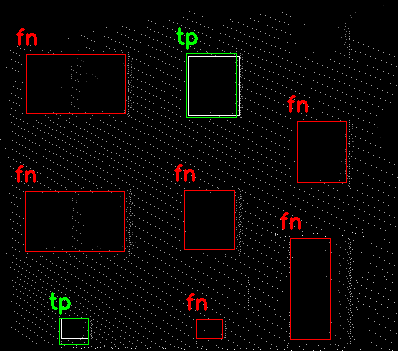}\vspace{0.5mm}\\
    \includegraphics[clip, trim=0 60 100 25,width=\textwidth]{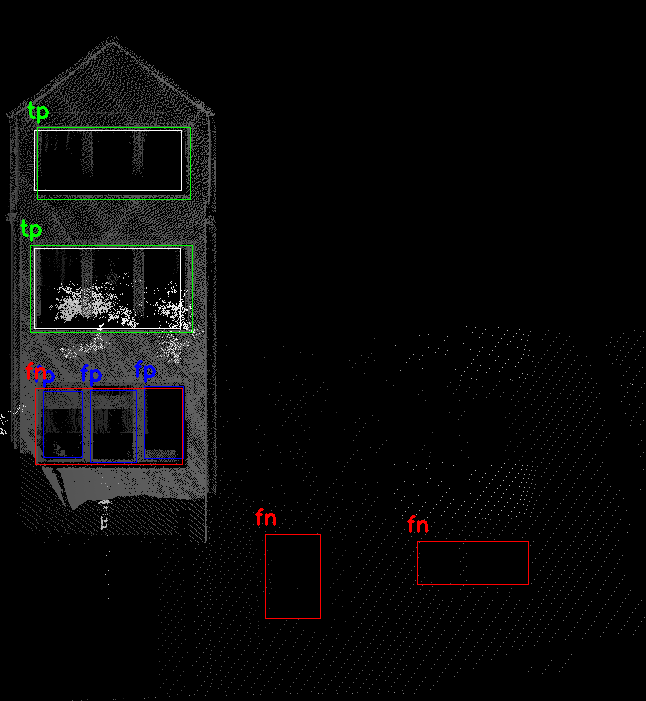}\\
    UNSUPERVISED\\
    \end{minipage}%
    }%
    \hspace{1mm}
  \subfigure{
    \begin{minipage}[t]{0.235\linewidth}
    \centering
    \includegraphics[clip, trim=410 20 20 20,width=\textwidth]{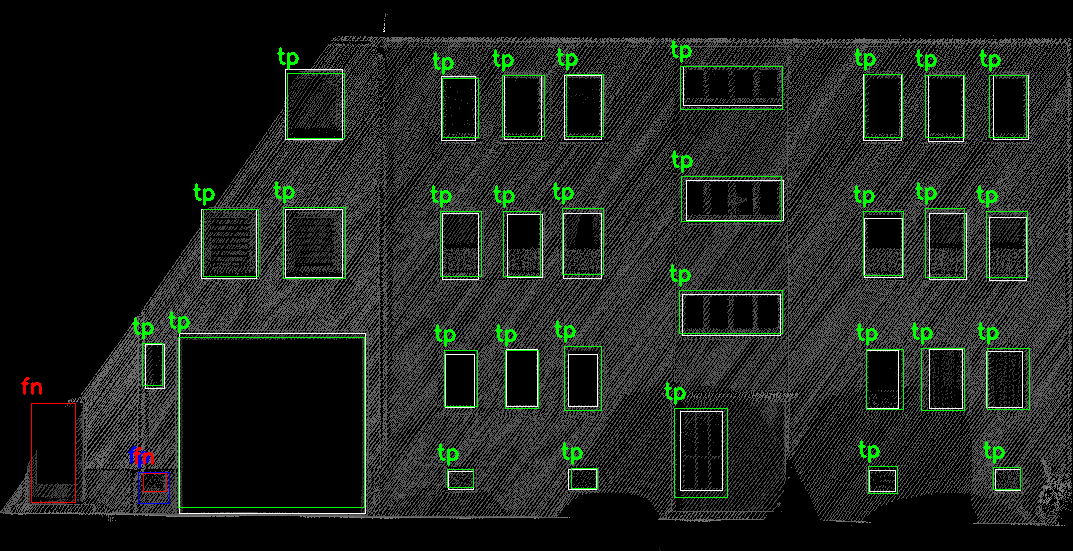}\vspace{0.5mm}\\
    \includegraphics[clip, trim=230 20 340 100,width=\textwidth]{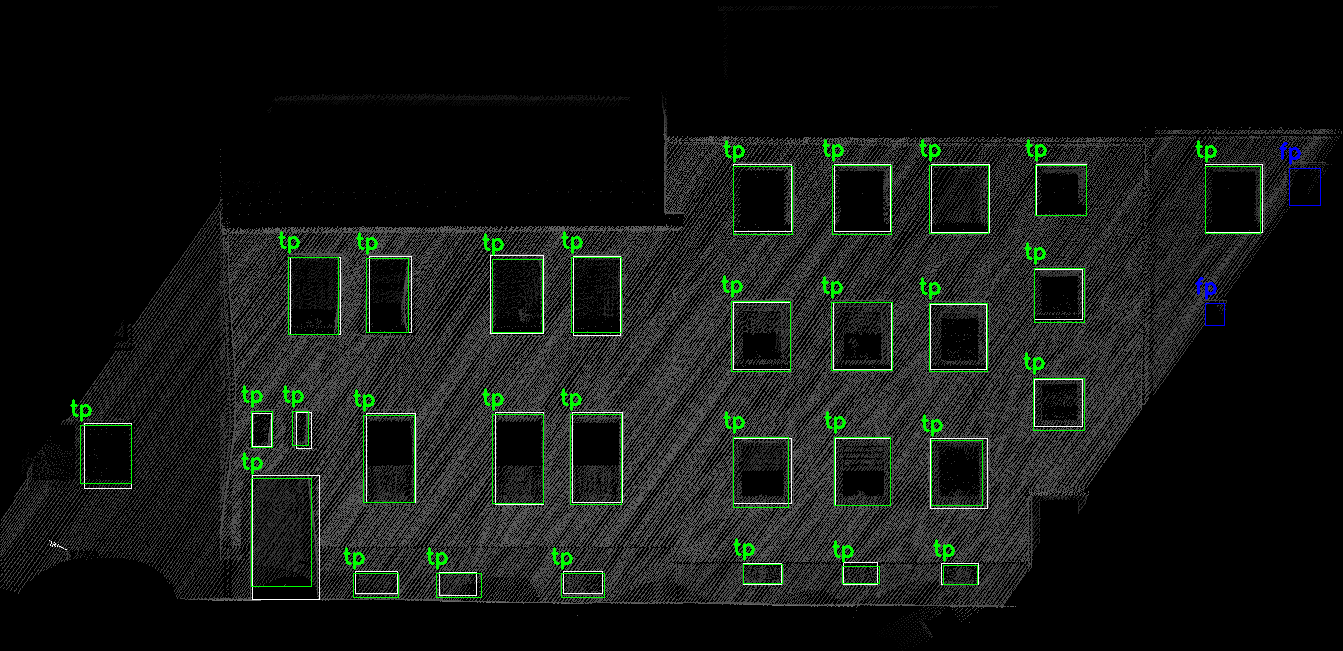}\vspace{0.5mm}\\
    \includegraphics[clip, trim=330 0 240 70,width=\textwidth]{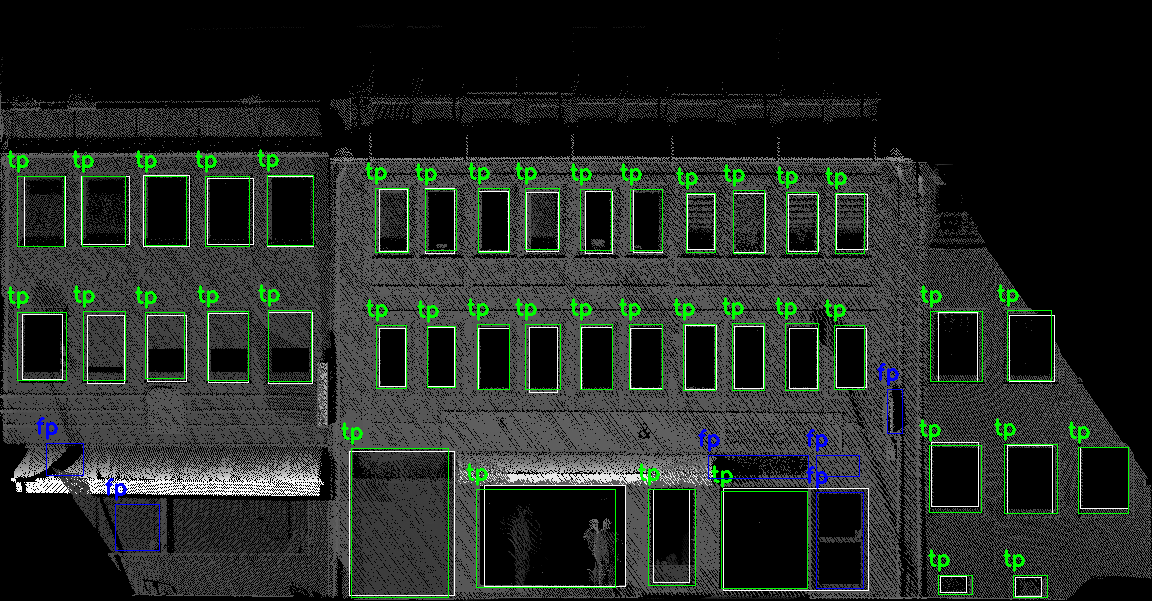}\vspace{0.5mm}\\
    \includegraphics[clip, trim=940 0 0 0,width=\textwidth]{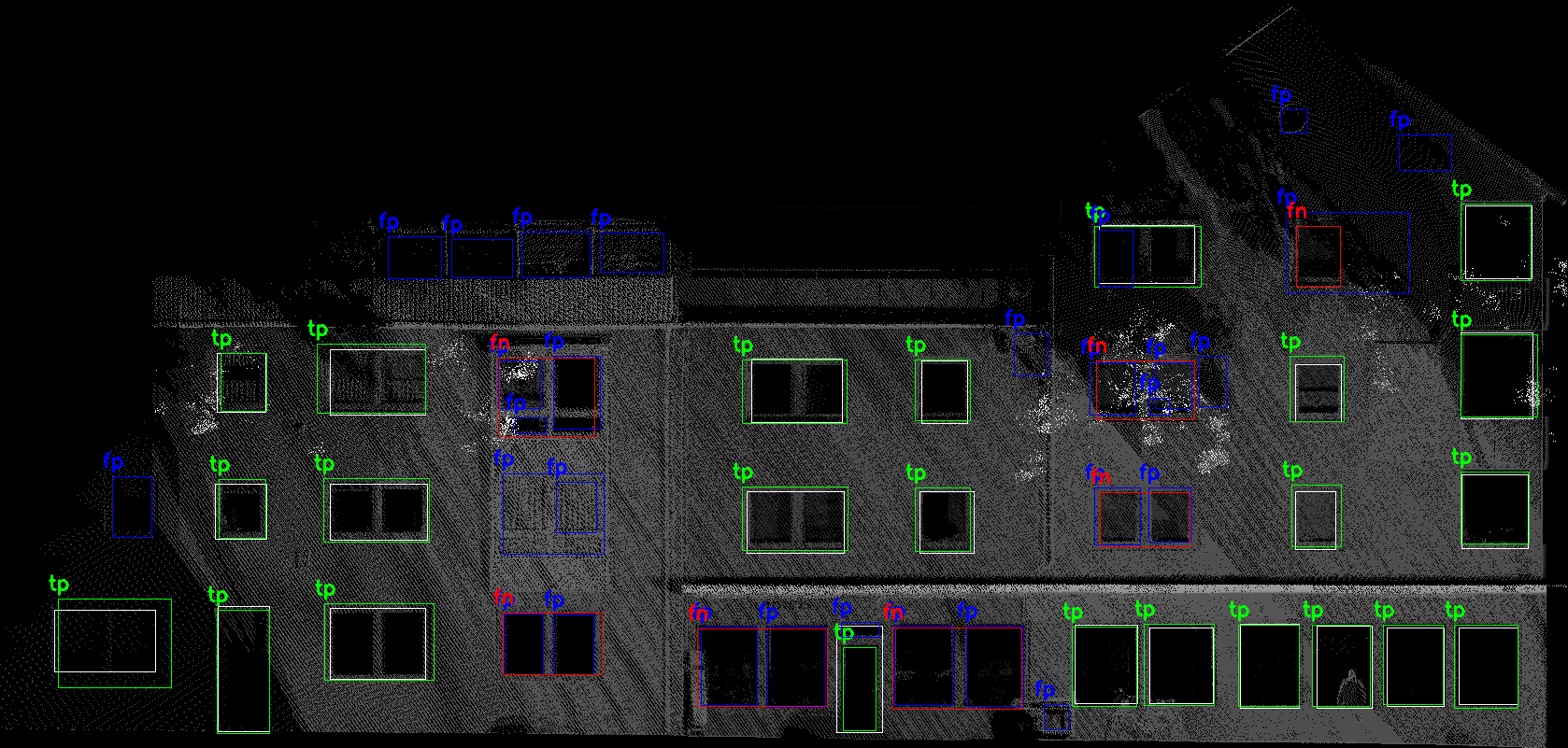}\vspace{0.5mm}\\
    \includegraphics[clip, trim=0 0 0 0,width=\textwidth]{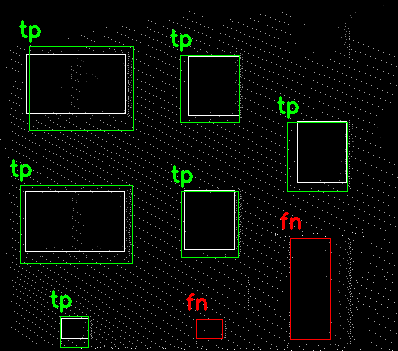}\vspace{0.5mm}\\
    \includegraphics[clip, trim=0 60 100 25,width=\textwidth]{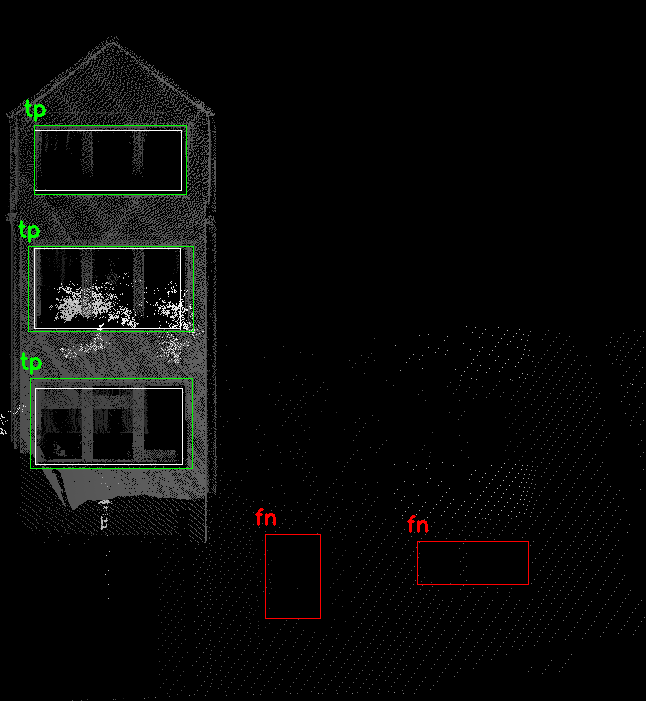}\\
    FILTERED\\
    \end{minipage}%
    }%
    \hspace{1mm}
  \subfigure{
    \begin{minipage}[t]{0.235\linewidth}
    \centering
    \includegraphics[clip, trim=410 20 20 20,width=\textwidth]{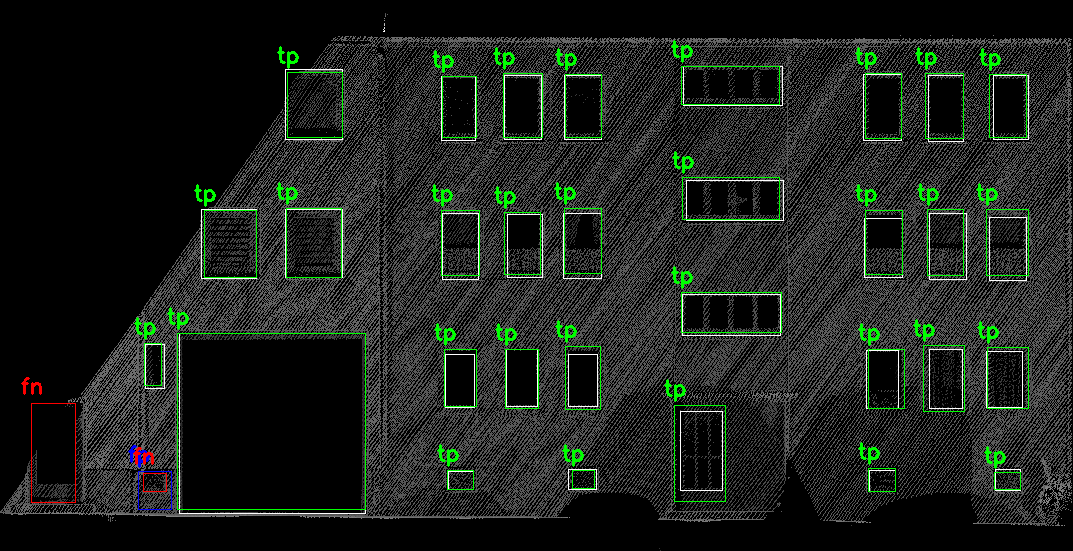}\vspace{0.5mm}\\
    \includegraphics[clip, trim=230 20 340 100,width=\textwidth]{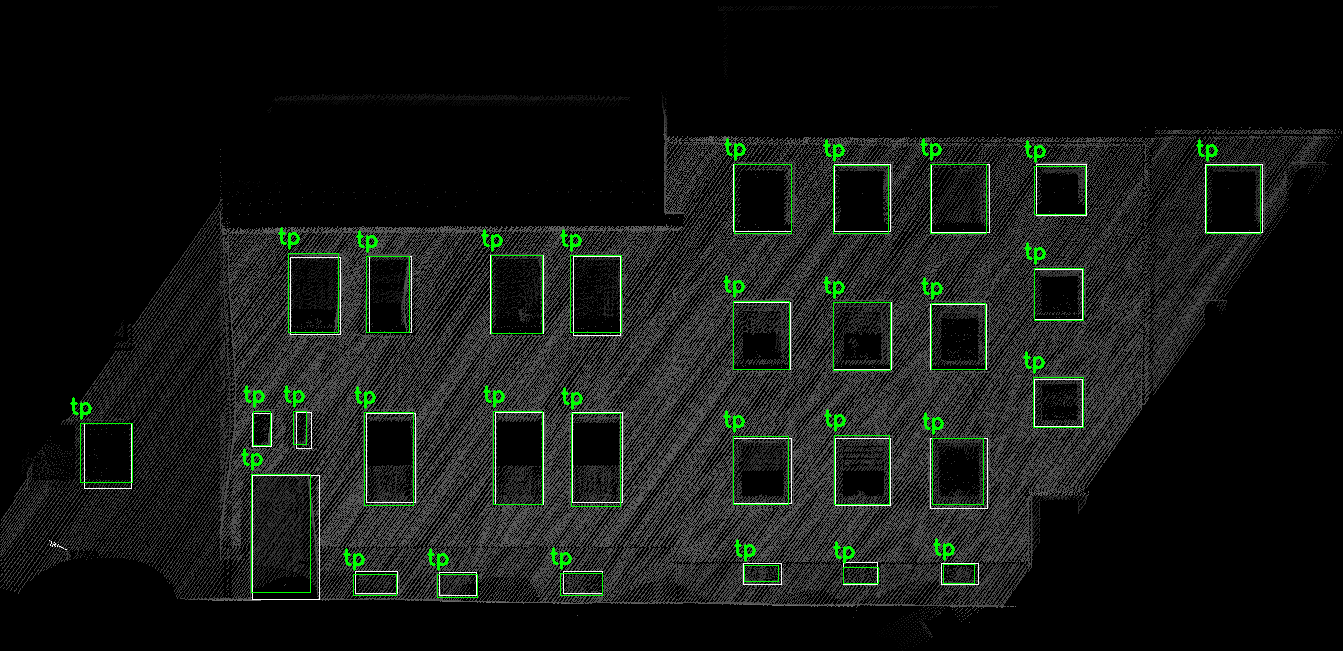}\vspace{0.5mm}\\
    \includegraphics[clip, trim=330 0 240 70,width=\textwidth]{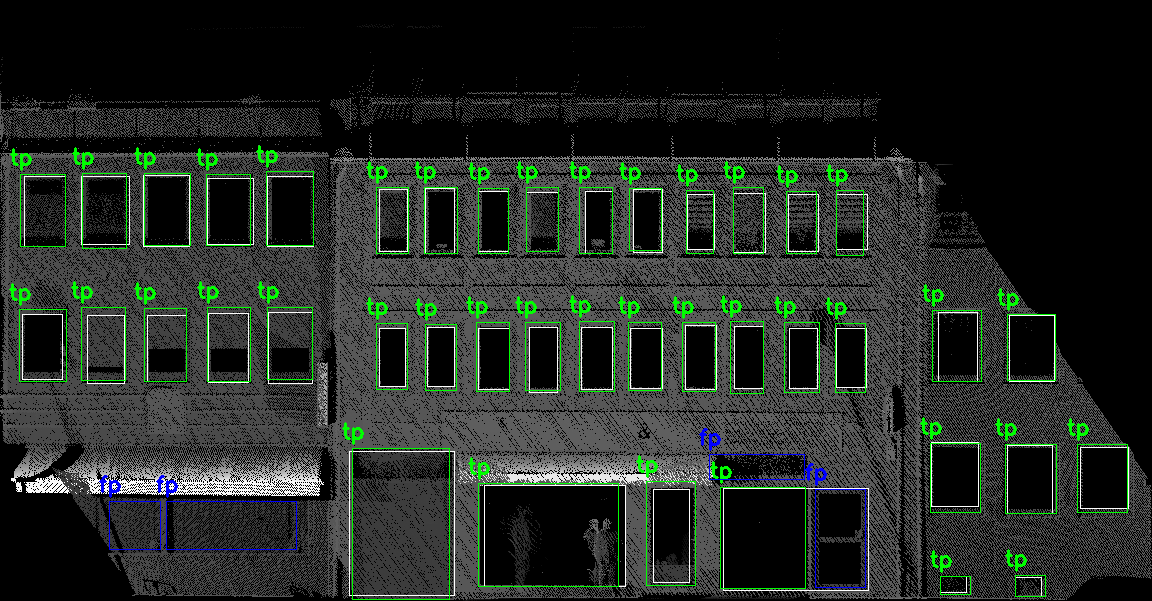}\vspace{0.5mm}\\
    \includegraphics[clip, trim=940 0 0 0,width=\textwidth]{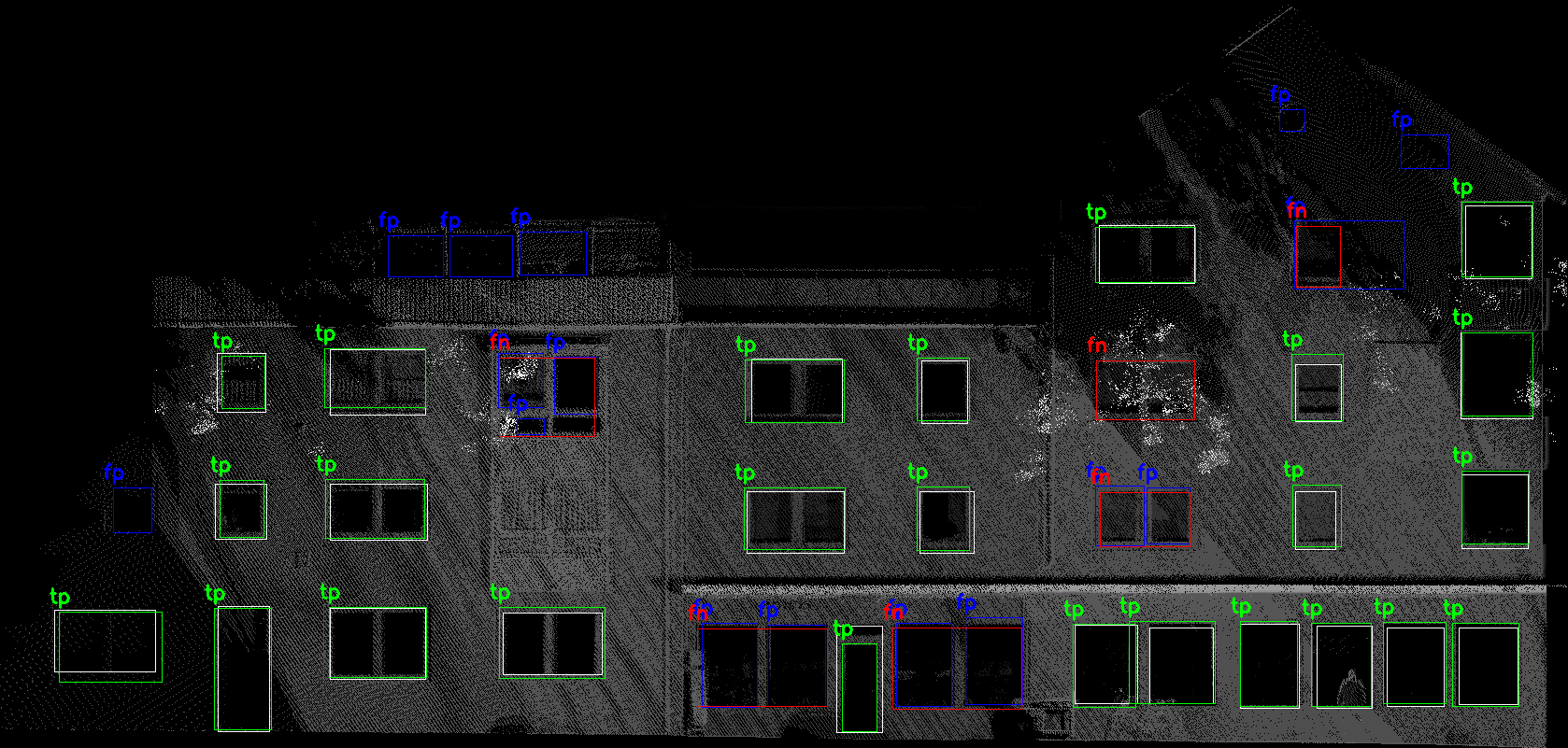}\vspace{0.5mm}\\
    \includegraphics[clip, trim=0 0 0 0,width=\textwidth]{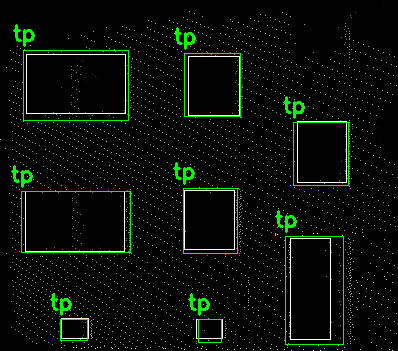}\vspace{0.5mm}\\
    \includegraphics[clip, trim=0 60 100 25,width=\textwidth]{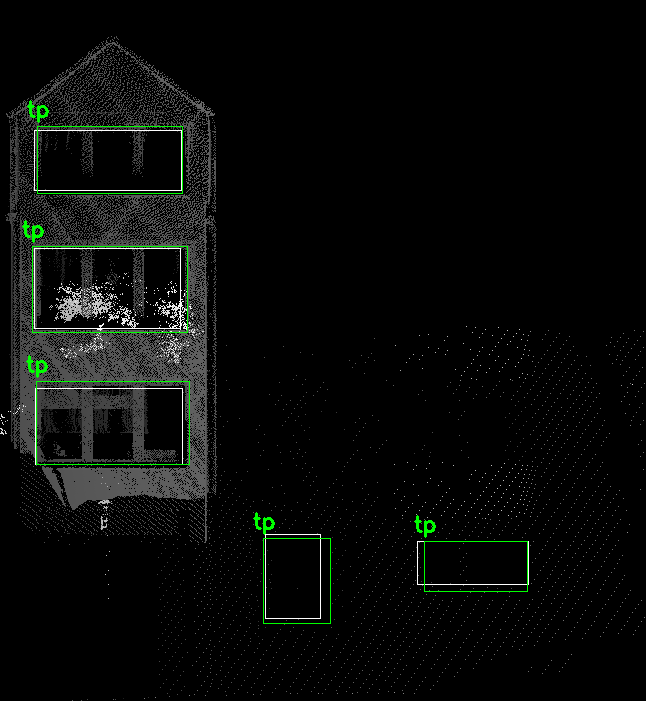}\\
    CORRECTED\\
    \end{minipage}%
    }
  \caption{Qualitative comparison of the detections using the rule-based method, and Faster R-CNN models trained on DATASET\_UNSUPERVISED, DATASET\_FILTERED, and DATASET\_CORRECTED.} 
  \label{fig:6_comp_good_example}
\end{figure*}


\begin{table*}
  \centering
  \renewcommand{\arraystretch}{1.3}
    \begin{tabular}{l l l c c c c c c}
    \hline
    Method & Image source & Dataset  & Precision & Recall & \Fone-score & $E_h$ [cm] & AVG [cm] & STD [cm] \\
    \hline
    Rule-based & Depth image & - & 0.645  & 0.654 & 0.650 & 6.03  & -0.53 & 9.75\\
    \hline
    Faster R-CNN & Composite image & UNSUPERVISED & 0.743 & 0.857 & 0.796 & 5.72 & 1.23 & 8.07\\
    (ResNet50+FPN & (depth, density, & FILTERED & 0.752 & 0.867 & 0.806 & 6.65 & 3.13 & 9.67\\
    as backbone) & reflectance) & CORRECTED & \textbf{0.825} & \textbf{0.893} & \textbf{0.858}  & \textbf{5.62} & 0.90 & \textbf{7.81}\\
    \hline
    Post processing  & & & & & & \\
    T1 & Composite image & CORRECTED & 0.900 & \textbf{0.931} & 0.916 & 5.73 & 0.88 & 8.41\\
    T1+T2 & Composite image & CORRECTED & \textbf{0.907} & 0.927 & \textbf{0.917} & 5.65 & 0.81 & 7.87\\ 
    \hline
    \end{tabular}%
  \caption{Comparison of detection results using the rule-based method and semi-supervised method. The backbone of Faster-RCNN in the semi-supervised method is ResNet50 + FPN.}
  \label{tab:comp_two_approaches}
\end{table*}



\begin{figure*}
  \centering
  \includegraphics[clip, trim=0 0 150 40, width=0.32\textwidth]{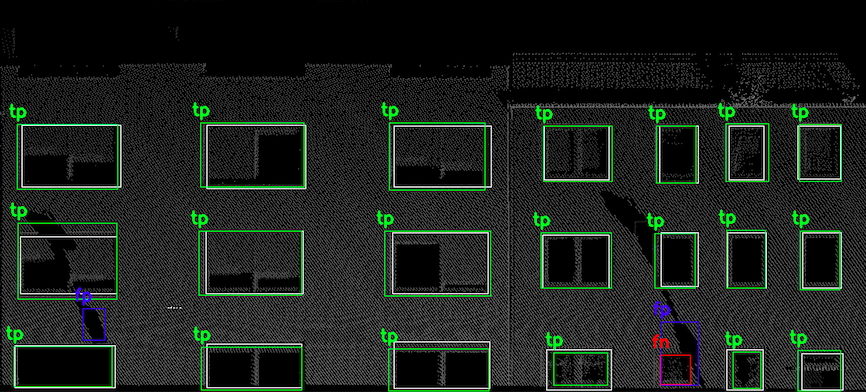}
  \includegraphics[clip, trim=0 0 150 40, width=0.32\textwidth]{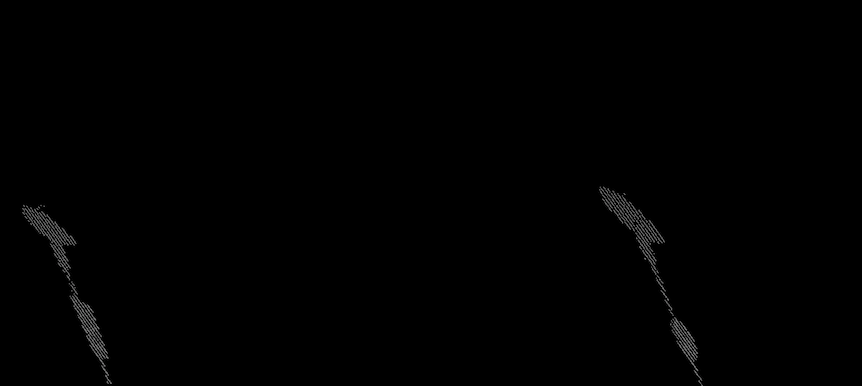}
  \includegraphics[clip, trim=0 0 150 40, width=0.32\textwidth]{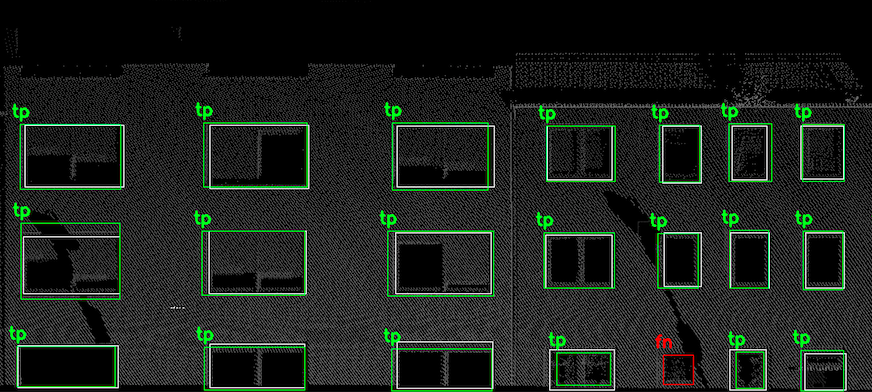}
  \caption{Example of the post-processing based on laser ray tracing. The prediction of the model (left) contains two false-positive predictions. An occlusion mask (middle) is calculated based on the laser ray tracing.
  According to the occlusion mask, some of the detections are rejected to provide the final output (right).}
  \label{fig:6_semi_post_occ_mask}
\end{figure*}

\begin{figure}
  \centering
  \includegraphics[width=0.23\textwidth]{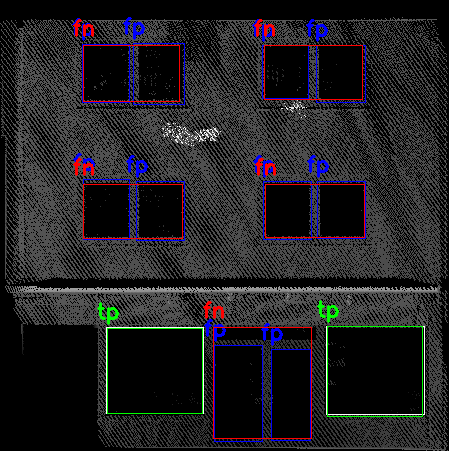}
  \includegraphics[width=0.23\textwidth]{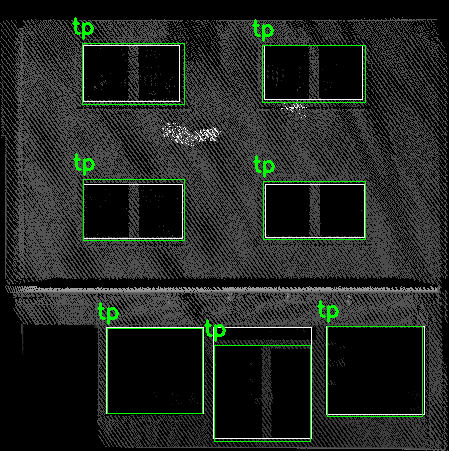}
  \caption{As for the purpose of our application scenario focusing on facade openings' height, the over-refined detections are not necessarily to be considered as false-positive detections. With more tolerance in the evaluation, the multiple detected objects within a ground truth box are merged. Shown are the evaluation results of the original model detection (left) and after merging inner boxes (right).}
  \label{fig:6_semi_post_1}
\end{figure}

\subsubsection*{Semi-supervised models for detecting windows and doors}

Although the rule-based method can achieve only a precision of
0.645, it already provides a large number of reasonable annotations.
The rule-based method is applied to all of the training facade images, i.e., the facades marked blue in Figure \ref{fig:overview_map}. They are used as the pseudo labels to train the Faster-RCNN model presented in Section \ref{sec:semi_supervised}. Due to the potentially high amount of label noise, different levels of supervision have been experimented with to train the same model. DATASET\_UNSUPERVISED contains pseudo labels without any additional supervision. 

It can be observed that many of the results from the rule-based model are less satisfying, especially for the areas with low point density. 
If these samples are involved in supervised learning, they may significantly interfere with the detection model.
Therefore, the images with many wrongly predicted pseudo labels are removed. 
It took around two hours' work for one annotator to inspect 2485 facade images, during which 838 images were removed. 
These removed images are mostly the facades with low point density or images without visible objects.
This dataset, obtained by low-level supervision, is named DATASET\_FILTERED.

Removing images with many incorrect pseudo-labels introduces fewer examples of errors during training. However, this does not provide new information for the learning of these undetectable samples.
Therefore, 165 representative images were selected out of the 838 removed images. 
The selected images are those of the facades facing the scanning vehicle and have a high point density. They should also include multiple detection targets in the same scene.
These images were manually annotated and added back into the training process.
The workload was around three hours for one annotator.
This dataset, obtained by significantly more human supervision, is named DATASET\_CORRECTED.

Faster-RCNN models with the same setting are trained based on the above-mentioned three datasets.
Transfer learning is applied. The ResNet50 backbone is first initialized with the weights pre-trained on the COCO dataset\footnote{COCO - Common Objects in Context. \url{https://cocodataset.org/} (Accessed on 08.03.2021)}. The early convolutional layers are frozen. Only the last three layers are fine-tuned on our dataset. This is a decision made after several trials training runs using multiple freeze combinations.

There are 143 extra annotated images used as the validation set. After observing the loss curve of the validation set, the maximum number of epochs was set to 20, as the model was no longer significantly optimized after 13 epochs in most of the trial training runs. The best model on the validation set was saved and applied in the follow-up prediction step. The training process usually takes around three hours using the Google Colab\footnote{Google Colaboratory. \url{https://colab.research.google.com/} (Accessed on 08.03.2021)} TPU environment.

The results are compared with the rule-based methods both qualitatively and quantitatively in Figure \ref{fig:6_comp_good_example} and Table \ref{tab:comp_two_approaches}.
The first three rows in the Figure \ref{fig:6_comp_good_example} present the comparison of models for the building facades with sufficient point density. Many facade openings are not detected by the rule-based method but are detected by the learning-based approach properly. The third row contains several false-positive detections. Some annotations are considered single targets, but the models detect their parts as small targets.
Sometimes the window spacing can be large or small. It is therefore difficult for the annotator to be perfectly consistent in these particular cases. 
Therefore, such false-positive detections should not necessarily be considered errors for our application scenario when more attention is paid to the bottom edges' height.

The last three rows in Figure \ref{fig:6_comp_good_example} present the performance of models for the building facades with relatively low point density. It can be observed that for the rule-based approach, many of the facade openings are not detectable (row 5), and the false-positive detections appear very frequently in the area of low point density (row 4). The learning-based method achieved a much better performance, especially after filtering the images with many wrong pseudo labels. After the correction, the learning-based model achieves much better results.

Table \ref{tab:comp_two_approaches} presents the comparison of the detection performance using different methods and post-processing techniques.
In addition to the detection metrics and averaged height error $E_h$, the mean (AVG) and standard deviation (STD) of the height errors compared to the targets are reported to demonstrate the distribution of height errors.
The learning-based method using this semi-supervised strategy without any further supervision has achieved a 14.6\% improvement on the \Fone-score over the rule-based method. With low-level supervision, i.e., filtering of the images with many false detections, the precision of the models improves slightly by 1\% on the \Fone-score.
After the correction, the model trained with data has achieved the best performance, which is 20.8\% over the rule-based method and 6.2\% over the learning-based model without additional supervision. The height errors of the bottom edge are also minimal.

As for the true-positive examples, the learning-based me\-thod, in general, achieves better results than the rule-based method on averaged height differences $E_h$. 
%
The height estimation accuracy (i.e., AVG) of the rule-based approach is slightly higher compared to the best of the learning-based approach, however, the difference is very small. The learning-based methods all have a gain on the precision of height estimation (i.e., STD), and the best learning-based method outperforms the rule-based approach by around 2~cm.

Furthermore, two types of post-processing are considered. As for our application scenario, the false-positive detections, as we observed in the third row of Figure \ref{fig:6_comp_good_example}, should not necessarily be regarded as errors. Therefore, the metrics are given with more tolerance, where the multiple detected objects within a ground truth box are merged as shown in Figure \ref{fig:6_semi_post_1}. With this, all detection metrics, precision, recall, and \Fone-score, are above 90\% (shown in row `T1' in Table \ref{tab:comp_two_approaches}). 
In addition, as described in Section \ref{sec:post_processing}, laser rays are traced to recover occluded facade areas as a binary mask. This mask is also used in post-processing to reject detections on that mask, as shown in the example presented in Figure \ref{fig:6_semi_post_occ_mask}. Table \ref{tab:comp_two_approaches} row `T1+T2' shows a slight improvement when using this technique, where the precision increase is 0.7\%.

\subsection{Building's flood risk mapping}

A high-resolution Digital Terrain Model (DTM) has been computed using the method proposed in \cite{feng2018enhancing}. 
Ground measurements were extracted from mobile mapping LiDAR data and merged with the DTM product provided by the local mapping agency. The result has a high completeness as well as a high resolution in the vicinity of roads.

\begin{figure}
\centering
\includegraphics[width=\linewidth]{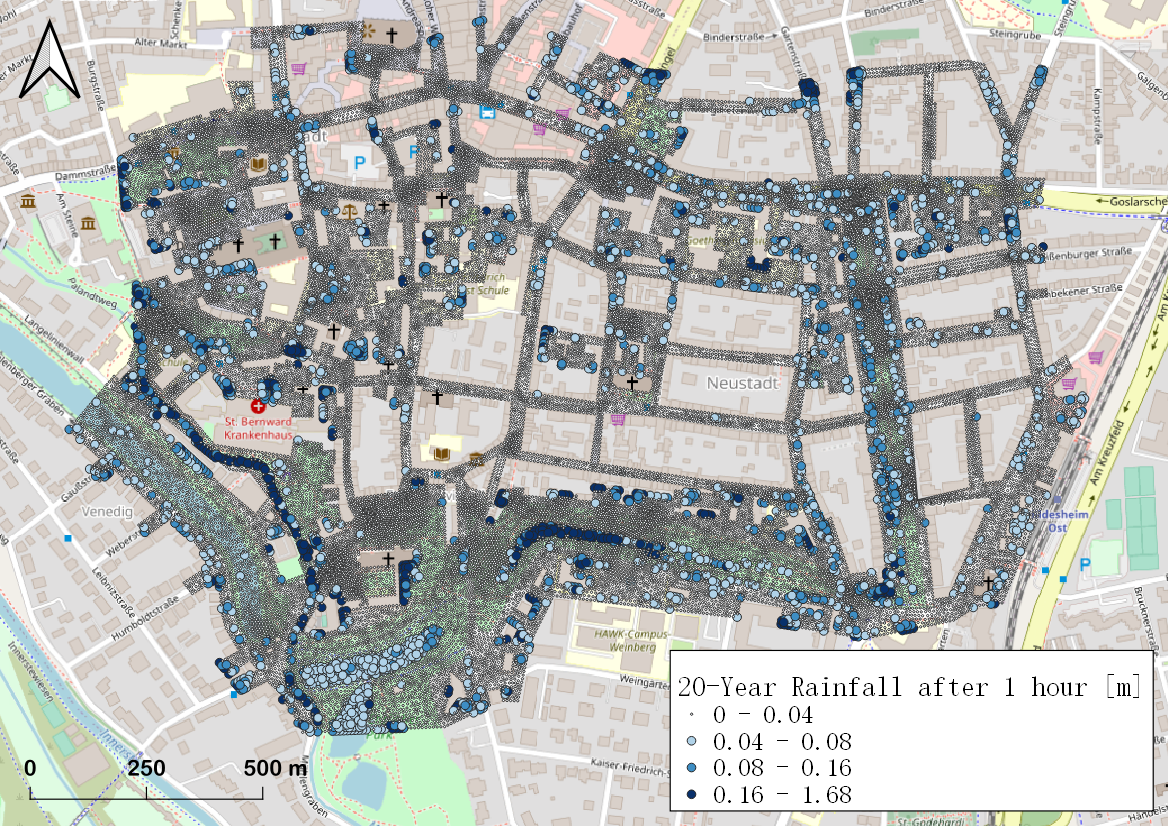}
\caption{Flood simulation results of a one-hour rainfall event with a return period of 20 years.}
\label{fig_sim}
\centering
\includegraphics[width=\linewidth]{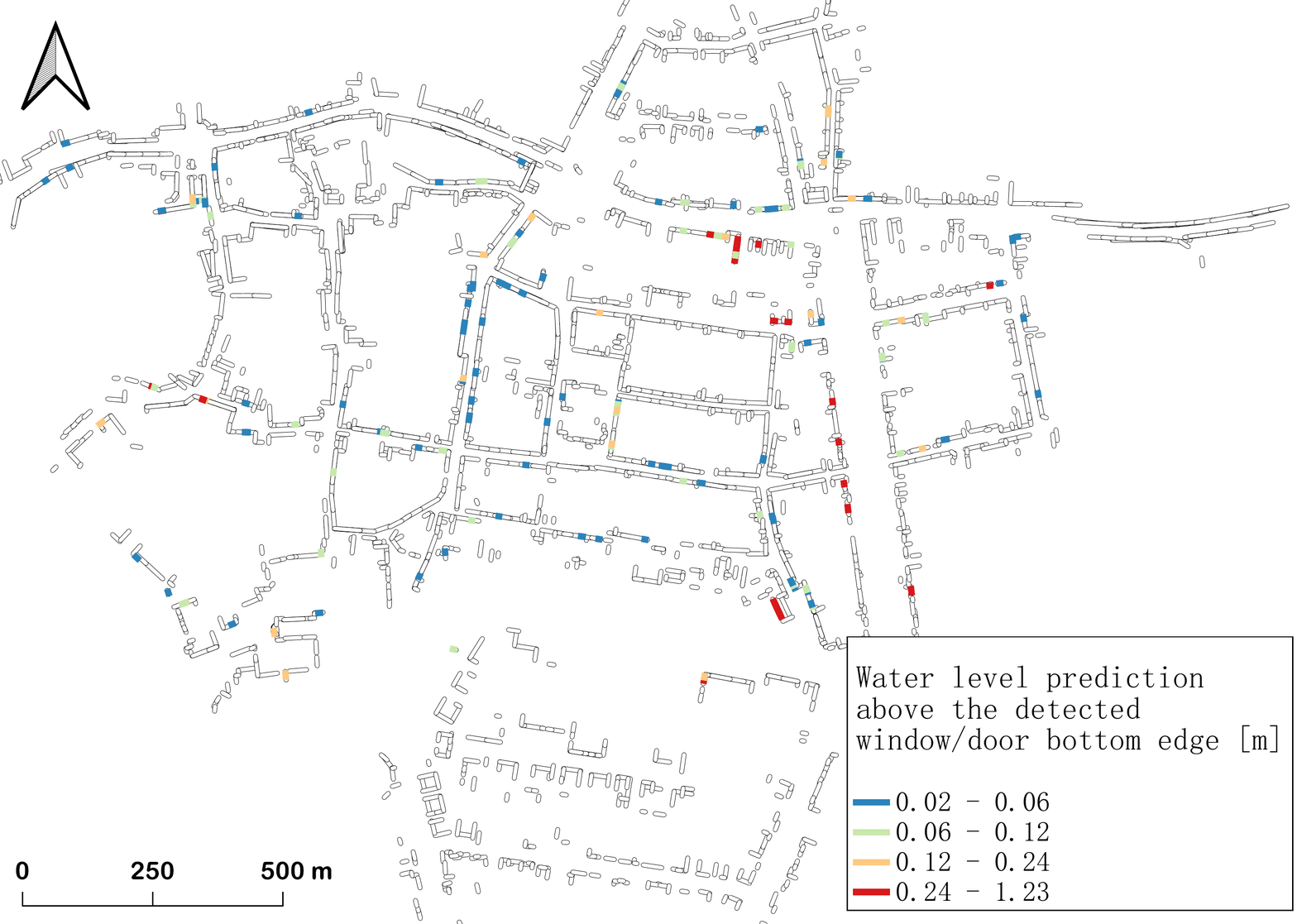}
\caption{Building flood risk map based on the simulation shown in Figure~\ref{fig_sim}.}
\label{fig_flood_risk_map}
\end{figure}

With this high-resolution DTM, a 2D hydraulic simulation of surface water runoff was performed using the software provided by BPI Hannover, as described in Section \ref{sec:indentification}.
Flood simulations are based on a statistical, region-specific rainfall event, i.e., a so-called Euler type II event \citep{wartalska2020analysis} for the region of the study area, of one-hour duration and with a return period of 20 years. The friction parameter to account for flow energy losses was chosen according to \citet{jankowski2021overland}.
The flood simulation results after one hour's rainfall are shown in Figure \ref{fig_sim}. 

\begin{figure*}
\centering
\includegraphics[clip, trim=0 0 240 0, width=\linewidth]{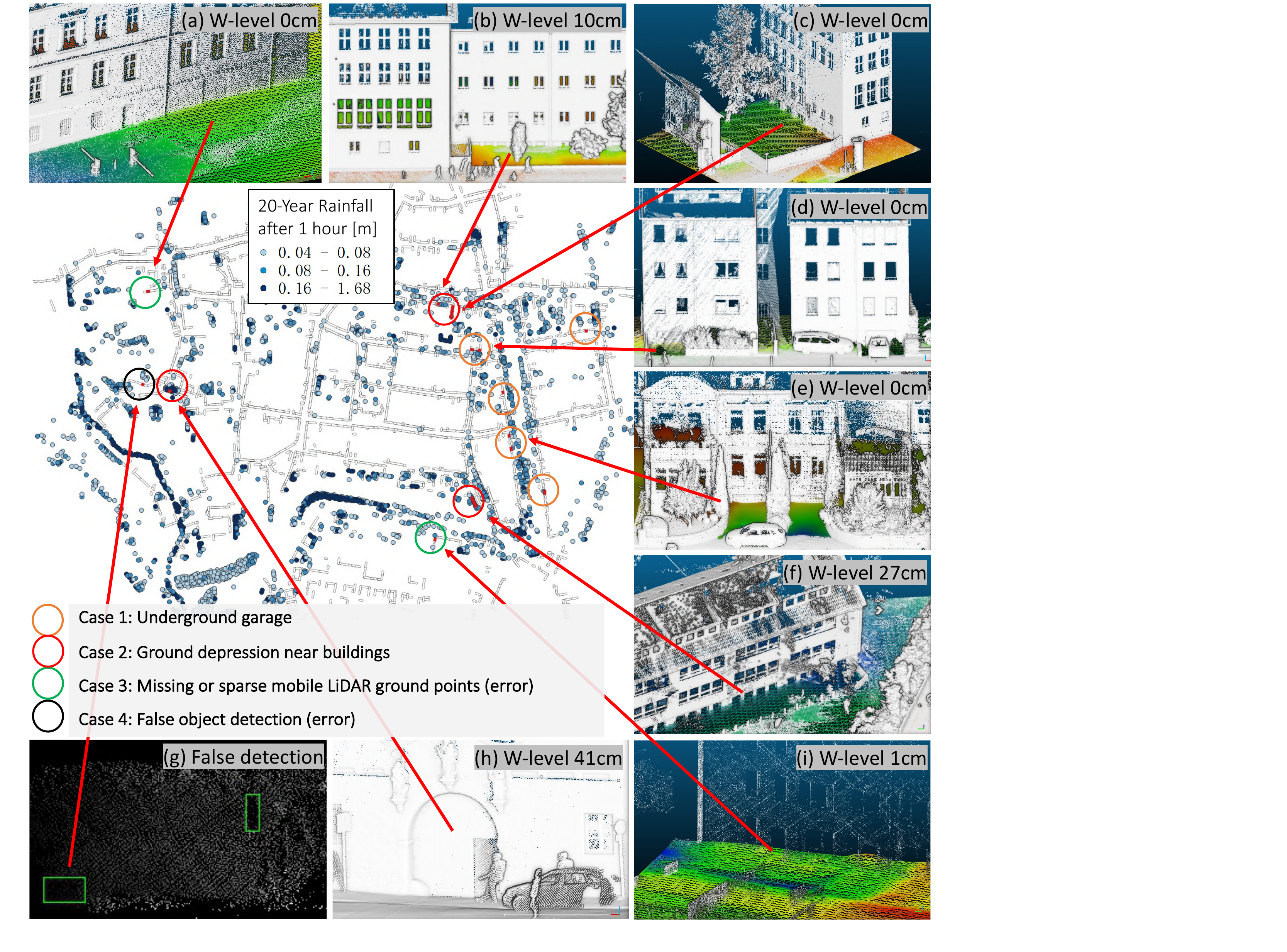}
\caption{Identified flood risk locations based on the simulation results shown in Figure~\ref{fig_sim} (red line: water above window/door bottom edge by more than 24~cm).}
\label{fig_risk_map_check}
\end{figure*}

We searched the nearest water level prediction data point for every detected window and door lower than two meters. 
The building flood risk value $I$ is then calculated for each detected line segment as the difference between the predicted water level and the detected bottom edge height above the DTM. 
Since the detection algorithm may contain errors of around 6~cm, the building flood risk map is visualized in Figure \ref{fig_flood_risk_map} with four risk levels. The higher these values are, the more likely it is that the line segments are exposed to flood risk. In order to check the identified building flood risk locations, the line segments with a flood risk index exceeding 24~cm were inspected individually for verification, as shown in Figure \ref{fig_risk_map_check}.

Four scenarios can be discerned. Case 1 includes the underground garages, which are indeed high-risk locations for buildings even though no flood water is predicted.
As the recessed area at the garage entrance often has fewer ground points than is necessary to update the DTM, the detected facade openings are substantially below ground level (Figure \ref{fig_risk_map_check}d and \ref{fig_risk_map_check}e), typically the underground garage entrance. 
Case 2 is another common risk location, where the ground is depressed near buildings (Figure \ref{fig_risk_map_check}c and \ref{fig_risk_map_check}f) or where windows are directly adjacent to or below the ground (Figure \ref{fig_risk_map_check}b). Figure \ref{fig_risk_map_check}h presents a gate where significant terrain depression exists, and a water level of 41~cm is predicted. Case 3 includes miss-identifications due to the imperfection of the reference DTM.
The mobile mapping LiDAR measurements are sometimes not perfectly matched with the airborne laser scanning produced terrain points due to unevenly distributed GNSS errors and the absence of ground measurement due to occlusions. As presented in Figure \ref{fig_risk_map_check}a and \ref{fig_risk_map_check}i, the detected doors from mobile mapping LiDAR data is significantly lower than the terrain provided by airborne laser scanning, even though there is no flood predicted at these two locations. Lastly, case 4 is caused by the false detection on the wall, as shown in Figure \ref{fig_risk_map_check}g.
As for the red edges, 44 out of 47 edges (13 out of 16 locations) are indeed exposed to flood risk.


\section{Discussion}

In this work, a pipeline is proposed to identify windows and doors at risk of water ingress. The facade openings are detected from mobile mapping LiDAR data, where the building facades are extracted and projected to 2D images for object detection. In order to detect the targets with high precision and with low annotation effort using deep learning models, a semi-supervised pipeline is applied, where first, pseudo labels are generated with a rule-based method using scan lines. 
With the simple rule-based approach it was difficult to achieve a good performance in all scenarios of the study area. However, the result was only needed as input for the deep learning method.
Deep learning models were then trained on the predictions using this rule-based method. The introduction of semi-supervised learning greatly improved the detection performance, and the additional supervision has significantly improved the robustness of models against very sparse point density.

Both the detection pipeline and the flood simulation models contain errors. The 2D images of facades are projected to 2~cm grids. Therefore, the derived heights of the facade openings are subject to this discretization error. Using smaller grids will result in a more sparse distribution of pixels, which may make the detection more difficult. 
Although the images are annotated as precisely as possible, the uncertainty of manual annotations exists for the 2~cm-resolution images in both the training data and the test data. 
Surveying some of the windows would provide the most accurate true values, but would require substantial extra fieldwork.
In addition, the evaluation gives equal importance to each individual facade opening. However, accurate detection of openings at low elevations should be considered as more important than detection of openings at high elevations. Therefore, further investigation is needed to design a metric that reasonably considers this difference in importance.

At the same time, it is worth noting that the flood simulation models also have errors (based on linearization of non-linear processes, parameterization, and choice of parameter set, temporal and spatial discretization). The validation and verification of flood simulation models is still an active research area, and the models used in the present study are subject to constant quality control and benchmarking \citep{kolditz2012thermo,maxwell2014surface}. Since flood hazards often occur quickly, it is difficult to make a reasonable estimation of the accuracy of study area-specific model results without sufficient on-site observations.

As for the mapping process, estimating the ground reference height of the detected facade openings is sometimes not accurate enough, as case 3 presented in Figures\ref{fig_risk_map_check}(a) and \ref{fig_risk_map_check}(i). This is due to the fact that the ground was sometimes not visible by the scanner or the ground measurements were too sparse to update the DTM. The current approach uses DTM products from local mapping agencies as an alternative for areas without mobile mapping LiDAR measurements. 
However, these DTM products are reported to have a height accuracy less than 30 cm\footnote{Digital terrain model - LGLN (German). \url{https://www.lgln.niedersachsen.de/startseite/geodaten_karten/3dgeobasisdaten/dgm/digitale-gelaendemodelle--dgm-143150.html} (Accessed on 08.03.2021)}.

Nevertheless, the incompleteness of the facades caused by obstacles such as low walls, fences, vehicles, and pedestrians, has always been the greatest challenge. This can lead to the missing and incompleteness of facade openings at lower parts of the facades. 
In the data used in this study, many facades have been measured several times, therefore, the occluding effect of dynamic objects (such as cars) could be reduced. Still the occlusions cased by walls, fences can only be complemented by additional measurements.
%
In addition, because of the limited accessible regions of mobile mapping vehicles, the building backs are mostly not visible and building sides are often scanned only with lower point-density.

A LiDAR mounted to an unmanned aerial vehicle (UAV) would be another commonly used device to obtain a large number of measurements at an oblique view angle. 
For places that are difficult to reach with vehicles and UAVs, backpack wearable mobile mapping can also be used.
It has a great chance to acquire occluded facade openings that are not accessible to terrestrial and mobile mapping LiDAR, which would be useful to complement the current pipeline. 
However, as of today, the mobile mapping LiDAR data acquisition is easier and faster compared to UAV-mounted LiDAR that is hampered by power supply and flight permission issues.

If no additional data acquisition is considered, on the methodological side general architectural knowledge about structures and symmetries could be used to derive possible openings on the non- or only partially visible building facades.




\section{Conclusion}

In this paper, we present a pipeline method to generate large-scale building flood risk maps. Windows, doors, and underground garage entrances are places with high flood risk for buildings. With mobile mapping LiDAR measurements, facades are extracted and projected as 2D images for object detection. These facade openings are first detected with a rule-based method using scan lines. Deep learning-based object detection models are then trained based on the pseudo labels from the rule-based model and combined with human supervision at different levels.

The results show that using only the automatically generated pseudo-labels, the learning-based model outperforms the rule-based method by 14.6\% in terms of \Fone-score. After 5 hours of manual supervision, including removing images with many wrong pseudo-labels and the re-labeling of some of the removed images, the model can be improved by another 6.2\%.
After post-processing, the trained detection model can achieve an \Fone-score of 92\% and a height error of 5.7~cm for the test dataset. The proposed method is a scalable and automatic approach, which can be easily applied once the mobile mapping data is available.

This research generates a new geographic information layer of the location and heights of facades openings in the city.
By combining these detections with the flood simulation results, windows and doors exposed to flood risk are identified by a simple layer overlay.
With this, targeted prevention measures can be applied to achieve a fine-grained emergency response.

As for future work, image-based detection can be used to validate the detections from LiDAR data and complement facade openings that are not identified. Targets that can be detected from both data sources would have a higher confidence. In addition to terrestrial data acquisition, LiDAR mounted on drones can be used as supplement information, especially in areas occluded by low walls and fences.
A survey of window and door heights in the test area may also be considered to obtain the most reliable ground truth heights for the comparison of model performance. 
In addition to the forecast flood water depth, water level estimations can also be extracted from geotagged social media images or surveillance camera data, based on people \citep{feng2020flood}, vehicles \citep{park2021computer}, or stop signs \citep{kharazi2021flood} in the flood water. This real-time information source has the potential to optimize the current forecast-based building flood risk mapping.

Furthermore, autonomous vehicles are starting to enter people's lives. Many of these cars are equipped with LiDAR sensors that measure their surroundings, including building facades as well. If these crowdsourced data can be reasonably integrated, it has the potential to provide such a building flood risk map easily available to all areas accessible by autonomous vehicles.

\section*{Acknowledgment}
The authors would like to acknowledge the support from the BMBF funded research project ``TransMiT -- Resource-optimized transformation of combined and separate drainage systems in existing quarters with high population pressure'' (BMBF, 033W105A).


\bibliographystyle{cas-model2-names}

\bibliography{cas-refs}

\end{document}